\documentclass[10pt,journal,compsoc]{IEEEtran}

\usepackage[pagebackref=false,breaklinks=true,colorlinks=true,citecolor=blue,bookmarks=false]{hyperref}
\usepackage{amsmath}
\usepackage{amssymb}
\usepackage{amsthm}
\usepackage{graphicx} 
\usepackage{booktabs}
\usepackage{algorithm}
\usepackage{algorithmic}
\usepackage{url}
\usepackage{xspace}
\usepackage{xcolor}
\usepackage{multirow}
\usepackage{color}
\usepackage{array}
\usepackage{pifont}
\usepackage{bm}
\usepackage{subfig}
\usepackage[accsupp]{axessibility}
\usepackage{diagbox}

\usepackage{cite}

\ifCLASSINFOpdf
\else
\fi

\begin{document}
%
% paper title
% Titles are generally capitalized except for words such as a, an, and, as,
% at, but, by, for, in, nor, of, on, or, the, to and up, which are usually
% not capitalized unless they are the first or last word of the title.
% Linebreaks \\ can be used within to get better formatting as desired.
% Do not put math or special symbols in the title.
\title{Unsupervised Part Discovery via Dual Representation Alignment}
%
%
% author names and IEEE memberships
% note positions of commas and nonbreaking spaces ( ~ ) LaTeX will not break
% a structure at a ~ so this keeps an author's name from being broken across
% two lines.
% use \thanks{} to gain access to the first footnote area
% a separate \thanks must be used for each paragraph as LaTeX2e's \thanks
% was not built to handle multiple paragraphs
%
%
%\IEEEcompsocitemizethanks is a special \thanks that produces the bulleted
% lists the Computer Society journals use for "first footnote" author
% affiliations. Use \IEEEcompsocthanksitem which works much like \item
% for each affiliation group. When not in compsoc mode,
% \IEEEcompsocitemizethanks becomes like \thanks and
% \IEEEcompsocthanksitem becomes a line break with idention. This
% facilitates dual compilation, although admittedly the differences in the
% desired content of \author between the different types of papers makes a
% one-size-fits-all approach a daunting prospect. For instance, compsoc 
% journal papers have the author affiliations above the "Manuscript
% received ..."  text while in non-compsoc journals this is reversed. Sigh.

\author{Jiahao Xia, Wenjian Huang, Min Xu, Jianguo Zhang, Haimin Zhang, Ziyu Sheng and Dong Xu % <-this % stops a space
\IEEEcompsocitemizethanks{
\IEEEcompsocthanksitem J. Xia, M. Xu, H. Zhang and Z. Sheng are with the Faculty of Engineering and IT, University of Technology Sydney, Ultimo, NSW, 2007, Australia (e-mail: Jiahao.Xia@student.uts.edu.au; Min.Xu@uts.edu.au; Haimin.Zhang@uts.edu.au; Ziyu.Sheng@student.uts.edu.au).\protect
\IEEEcompsocthanksitem W. Huang and J. Zhang are with the Department of Computer Science and Engineering, Southern University of Science and Technology, Guangdong 518055, China (e-mail: huangwj@sustech.edu.cn; zhangjg@sustech.edu.cn). \protect
\IEEEcompsocthanksitem X. Dong is with the Department of Computer Science, The University of Hong Kong, Hongkong. (e-mail: dongxu@hku.hk). \protect
\IEEEcompsocthanksitem Corresponding author: Min Xu} \protect
\thanks{This work was supported by the program of China Scholarships Council under Grant 202006130004.}
% <-this % stops an unwanted space
}

\IEEEtitleabstractindextext{%
\begin{abstract}

% Finally, the geometric and semantic constraints provided by a novel representation transfer module and a CNN-based decoder are applied 

Object parts serve as crucial intermediate representations in various downstream tasks, but part-level representation learning still has not received as much attention as other vision tasks. Previous research has established that Vision Transformer can learn instance-level attention without labels, extracting high-quality instance-level representations for boosting downstream tasks. In this paper, we achieve unsupervised part-specific attention learning using a novel paradigm and further employ the part representations to improve part discovery performance. Specifically, paired images are generated from the same image with different geometric transformations, and multiple part representations are extracted from these paired images using a novel module, named PartFormer. These part representations from the paired images are then exchanged to improve geometric transformation invariance. Subsequently, the part representations are aligned with the feature map extracted by a feature map encoder, achieving high similarity with the pixel representations of the corresponding part regions and low similarity in irrelevant regions. Finally, the geometric and semantic constraints are applied to the part representations through the intermediate results in alignment for part-specific attention learning, encouraging the PartFormer to focus locally and the part representations to explicitly include the information of the corresponding parts. Moreover, the aligned part representations can further serve as a series of reliable detectors in the testing phase, predicting pixel masks for part discovery. Extensive experiments are carried out on four widely used datasets, and our results demonstrate that the proposed method achieves competitive performance and robustness due to its part-specific attention. The codes and models are available at https://github.com/Jiahao-UTS/UnsupervisedPart.

\end{abstract}
% The facial landmarks predicted by existing methods have unique correspondence to the pre-defined landmarks of the training dataset.
% Note that keywords are not normally used for peerreview papers.
}

\maketitle

\IEEEdisplaynontitleabstractindextext

\IEEEpeerreviewmaketitle

\IEEEraisesectionheading{\section{Introduction}\label{sec:introduction}}

\IEEEPARstart{O}{bject} parts play an important role in both human cognition and object interactions. In the realm of computer vision, object parts also serve as effective intermediate features for various downstream tasks, such as face editing \cite{MASKGAN}, human pose estimation \cite{Human_Pose_Part, DensePose} and 3D object reconstruction \cite{Construction3D}. However, annotating object parts is both costly and time-consuming, making it difficult for existing models to learn about object parts in a supervised manner. As a result, unsupervised part discovery has gained increased attention from the community in recent years.

Previous works \cite{DINO} have shown that vision transformer (ViT) \cite{VIT} can learn an instance-level attention without labels, which explicitly includes scene layout and object boundaries, and this attention can be directly applied to foreground object segmentation. Furthermore, the interpretability of the attention map also has a very close connection with the quality of the ViT features \cite{DINO_Reg}. As a result, these enhanced ViT features can also be leveraged to boost the performance of clustering-based unsupervised foreground/background segmentation \cite{LOST}, \cite{NormCut}, \cite{TokenCut}, \cite{DINO_vote}, \cite{DINO_background}, \cite{DINO_high}. Inspired by these works, we envision that this attention mechanism can also be utilized to improve the performance of unsupervised part discovery. However, unsupervised part discovery requires part-level attention while the instance-level attention is not fine-grained enough. Learning part-specific attention is particularly challenging because both supervised and self-supervised ViT commonly produces features with high similarities for the different parts of the same categories, as evidenced in \cite{amir2021deep}. These similar features within the same category make it difficult to distinguish a single part from entire target object to form part-level attention. Despite its importance, the challenge of encouraging ViT to learn part-specific attention in an unsupervised manner has not been adequately addressed.

\begin{figure*}[t!]
	\centering
	\includegraphics[width=\linewidth]{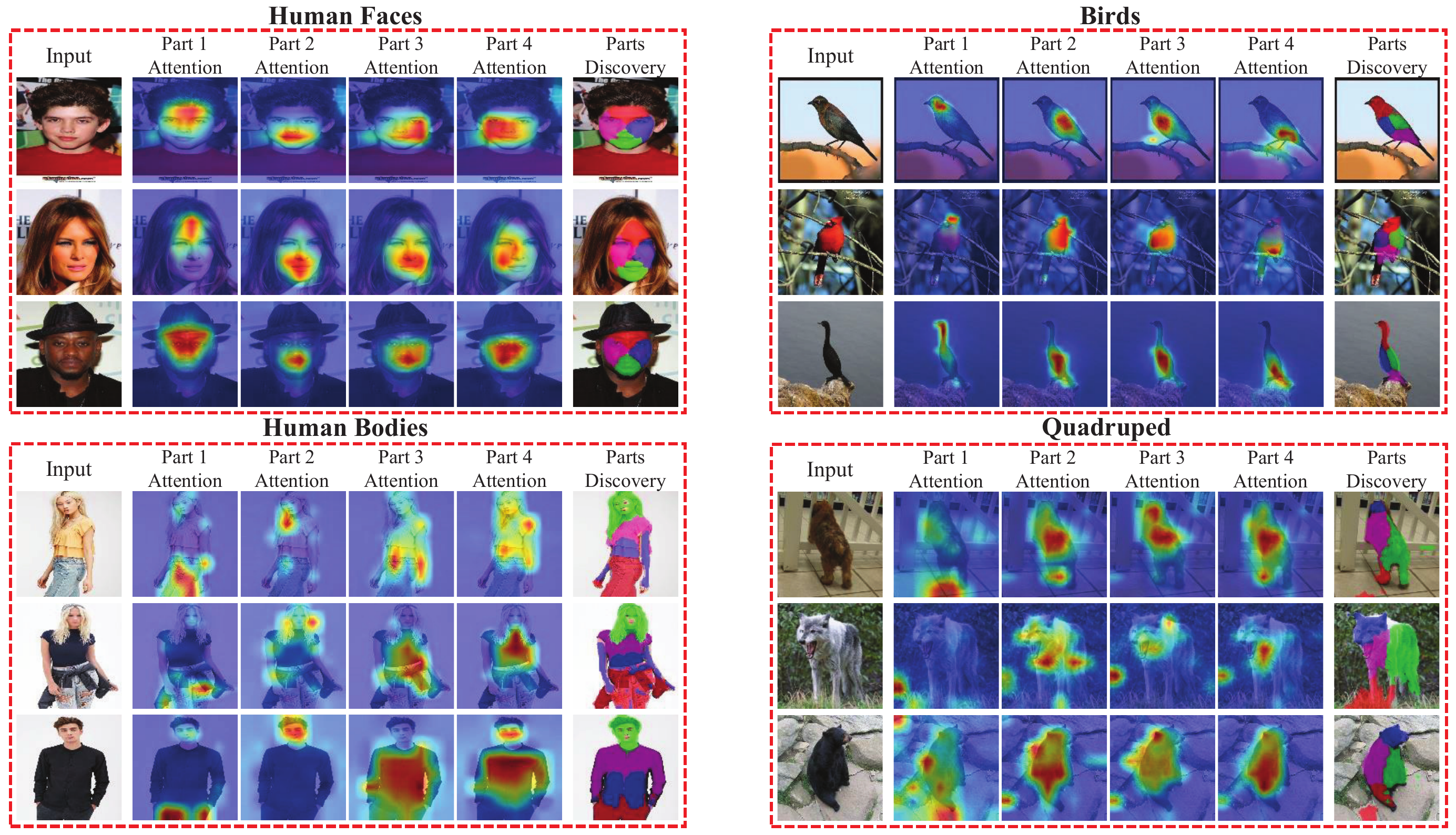}
	\caption{the self-attention maps to each object part and the unsupervised parts discovery results. The attention map is the self-attention averaged over the heads and layers, indicating the attention of each part embedding to image patch embeddings. The \textcolor[rgb]{1, 0, 0}{\textbf{red}}, \textcolor[rgb]{0, 1, 0}{\textbf{green}}, \textcolor[rgb]{0, 0, 1}{\textbf{blue}} and \textcolor[rgb]{1, 0, 1}{\textbf{pink}} area represent the part 1, part 2, part 3 and part 4 respectively.}
	\label{fig1}
\end{figure*}

% Intuitively, we believe the part-specific attention mechanism learning should satisfy at least two types of constraints

To drive the model to learn part-specific attention in an unsupervised manner, we employ two types of constraints to guide the learning process: \textit{semantic constraints} and \textit{geometric constraints}. The semantic constraints ensure the part representations explicitly contain the information of corresponding parts. Moreover, distinguishing the features of different parts within the same object forms the foundation for part-specific attention learning. Therefore, the semantic constraints also encourage the representations of different parts within the same object to be orthonormal, and the same predicted part across different targets share similar semantics. The geometric constraints guide the attention mechanism to focus on specific local regions of a target, as opposed to the entire image or a broadly salient object, from a geometric perspective. By combining semantic constraints with geometric constraints, the model can learn the part-specific attention.

Nevertheless, it is quite difficult to directly apply semantic and geometric constraints to part representations because they are a series of sparse vectors without explicit spatial information. To tackle this problem, we propose a novel dual representation alignment framework, which drives the part representations to have high similarity with the pixel representations of the corresponding regions and low similarity with those of the irrelevant regions in a dense feature map. Consequently, the geometric and semantic constraints are further applied to part representations through the intermediate results in dual representation alignment. Additionally, the alignment also enables the part representations to serve as a series of reliable detectors, predicting pixel-level masks during testing phase.

Specifically, we propose a PartFormer, replacing the learnable class embedding of a standard ViT with multiple learnable part embeddings for part representation extraction. Similar to \cite{DETR}, each part embedding learns different weights from training, which drives each part embedding to response to a specific part of the target object. Moreover, an additional semantic consistency constraint further encourages the different part representations to be orthonormal and the representations for the same part across different objects to be consistent. The part representations are then matched to a feature map with pixel representations extracted from the same image, producing a confidence map that indicates the semantic similarity between the pixel and part representations. The part representations are further assigned to form a dense feature map according to the confidence map. To align the part and pixel representations, we apply a concentration constraint on the confidence map, driving the part representations to have higher similarity with the pixel representations in the corresponding regions and lower similarity with those in other regions. This alignment allows each region on the synthetic feature map to be primarily represented by a single part representation. By applying a semantic constraint (reconstruction constraint) to this feature map, the majority of information from each part region is captured in its corresponding part representation. Moreover, the geometric constraints (concentration and area constraints) are also applied to PartFormer through the confidence map, encouraging it to focus locally.

To further the consistency of part discovery results, we extend the proposed framework to learn part representations with better geometric transformation invariance from paired images, which are generated from the same source image via geometric transformation. Then, they are fed into the PartFormer and their corresponding part representations are exchanged. Finally, the part representations and pixel representations from the pair images are used for producing the confidence map and reconstruction. This exchange encourages the part representations extracted from the paired images to be consistency, thereby leading to more consistent predicted results.

The proposed method is further validated on five widely-used datasets, including CelebA \cite{CelebA}, AFLW \cite{AFLW}, CUB \cite{CUB}, DeepFashion \cite{Deepfashion}, and PartImageNet \cite{PartImageNet}. Without relying on any auxiliary supervision, this proposed method is able to identify object parts from diverse backgrounds and outperforms the state-of-the-art methods. Additional \textit{cross}-dataset validation demonstrates that the proposed method has excellent generalization ability. Moreover, when we swap the part representations between two different images, the parts in the reconstructed images are also swapped according to the representations. These results illustrate that the part representations indeed capture the information of the corresponding parts.

The contributions of this paper are threefold:
\begin{itemize}
	\item To learn the part representations with high quality, we propose a novel paradigm that encourages the proposed PartFormer to learn part-specific attention in an unsupervised manner, using geometric and semantic constraints applied through dual representation alignment. The alignment also evolves the part representations into reliable detectors to further boost the performance in part discovery.

	\item To improve the consistency of the discovered parts, we further extend the dual representation alignment framework to paired images. By exchanging the part representations between paired images and using them for alignment and reconstruction, the learned part representations achieve better geometric transformation invariance, predicting more consistent part masks.
	
	\item Extensive experiments on unsupervised part discovery demonstrate the high-quality of the part representations learned by the proposed method. Moreover, the ablation studies comprehensively validate the effectiveness of the geometric and semantic constraints applied during training.
	
\end{itemize}

\section{Related Work}\label{sec2}

\subsection{Unsupervised/Self-supervised Learning}
Unsupervised/Self-supervised learning aims to learn effective representations without manual annotations using pretext tasks. Many current methods demonstrate that the image encoder trained by predicting image augmentations \cite{unsupervisedrotation}, \cite{unsuperviseddiscriminator}, \cite{unsuperviseddata}, discriminating between images \cite{MOCOV1}, \cite{MoCoV3} or clustering images with shared concepts \cite{Swav}, \cite{BYOL} performs better in downstream tasks than the model pretrained in a supervised manner. By further introducing certain components into self-supervised learning, such as momentum encoder and multi-crop augmentation, DINO \cite{DINO} produces the features that could achieve high accuracy on ImageNet \cite{ImageNet} using a basic nearest neighbors classifier without fine-tuning. To further encourage the pretrained model to produce more fine-grained features with part-level representations, Ziegler et al. \cite{Ziegler_self} and Saha et al. \cite{PARTICLE} extend clustering loss and contrastive loss to discriminate between pixels. Inspired by the success of Unsupervised/Self-supervised learning in natural language processing \cite{BERT}, several methods \cite{MST}, \cite{MAE}, \cite{SMA} also set recovering missing patches as the pretext task for representation learning. Nevertheless, only a small number of unsupervised/self-supervised ViT explicitly includes scene layout and object boundaries. Recent research \cite{DINO_Reg} finds that the interpretability of the attention map means less outlier tokens and higher feature quality. By encouraging the attention map to explicitly include scene layout and object boundaries, the performance of downstream tasks, such as unsupervised foreground segmentation, can be boosted significantly.

\subsection{Unsupervised Foreground Segmentation}
Unsupervised foreground segmentation aims to identify the foreground object region within an image. Early works in this field often relay on prior knowledge, such as color contrast \cite{GlobalContrast} and background prior \cite{BackgroundPrior} to achieve foreground segmentation. However, prior knowledge does not always yield reliable results in the presence of complex backgrounds, leading to fragile robustness. Croitoru et al. \cite{IJCVUnsuper} employ the VideoPCA algorithm \cite{VideoPCA} as the teacher network to train the student network for unsupervised foreground segmentation. However, the runtime of VideoPCA (4 seconds per frame) makes it impractical for use on large-scale datasets. To ensure training efficiency, they only run a part of VideoPCA system. However, the reduced accuracy of the teacher network introduces noise into the student network, leading to performance degradation. In recent years, generative adversarial network (GAN) based methods have come to dominate the filed of unsupervised foreground segmentation. Bielski et al. \cite{EmergenceObje} encourage GAN to disentangle background and foreground by introducing randomly generated and translated foregrounds and masks; Chen et al. \cite{ReDo} employ GAN to segment foreground masks by redrawing objects without changing the distribution of the dataset; Savarese et al. \cite{IEM} cluster foreground and background pixels from an information-theoretic perspective and generate pseudo-labels to train a segmentation model; Voynov et al. \cite{LargeScaleGenerativeModels} demonstrate that pretraining GAN on large-scale datasets can improve the object segmentation performance. Although GAN based methods significantly boost performance in unsupervised foreground segmentation, they still have drawbacks, such as falling into trivial solutions easily and not being trainable end-to-end. 

Caron et al. \cite{DINO} have proven that ViT can learn instance-level attention on large-scale datasets without labels, which offers a new direction for foreground object segmentation. With explicit scene layout and object boundary, the attention map can be directly used for segmentation. Furthermore, the interpretability of the attention map also has a very close connection with the quality of the ViT features \cite{DINO_Reg}. Therefore, the high quality features in~\cite{DINO} can also boost the performance of clustering-based unsupervised foreground segmentation~\cite{LOST},~\cite{NormCut},~\cite{TokenCut},~\cite{DINO_vote},~\cite{DINO_background},~\cite{DINO_high}. LOST~\cite{LOST} selects a foreground seed from the feature map, while Found \cite{DINO_background} selects a seed for background. They further expand the seed into a foreground or background mask based on the similarity of pixel features. Without selecting a seed, TokenCut~\cite{NormCut},~\cite{TokenCut} and Shin et al.~\cite{DINO_vote} construct a graph based on the feature similarity and directly partition this graph into two disjoinnt sets for foreground segmentation using spectral clustering. To predict more fine-grained masks, Ravindran et al.~\cite{DINO_high} employ a coarse-to-fine framework to predict the optimized masks with higher resolution from the coarse masks predicted by spectral clustering. However, without high quality feature map, the performance these clustering-based methods degrades significantly, as reported in \cite{DINO_Reg}. Therefore, learning high quality features is also a basic challenging in unsupervised foreground segmentation. 

\subsection{Unsupervised Part Discovery}
Unsupervised part discovery aims to decompose foreground objects into parts and predict pixel-level part masks. By introducing the part-level concept into networks, the performance of object classification \cite{HuangAttention}, \cite{PDiscoNet} and person re-identification \cite{Part_ReID} is improved. In the early stage, related works mainly focuses on unsupervised landmark detection \cite{ULD}, \cite{ZhangLandmark}, \cite{LOR_LANDMARK}, \cite{JakabLandmark}, \cite{AFLW2}, \cite{JakabVideo}, \cite{TejasLandmark}, \cite{KimLandmark}, \cite{MatthiasLandmark}, \cite{AliaksandrLandmark}, \cite{TUSK}, which is a less challenging task but has tight connections with unsupervised part discovery, because object landmarks can be viewed as a simplified form of part-level annotation.  Some methods for unsupervised landmark detection \cite{ULD, ZhangLandmark} can also predict pixel-level part masks, and some unsupervised part discovery methods \cite{LiuDis, SCOPS} are inspired by unsupervised landmark detection. Nevertheless, object landmarks are insufficient to describe object parts in some downstream tasks \cite{MASKGAN, Construction3D}. As a result, an increasing number of works are focusing on unsupervised part discovery.

\begin{figure*}[t!]
	\centering
	\includegraphics[width=\linewidth]{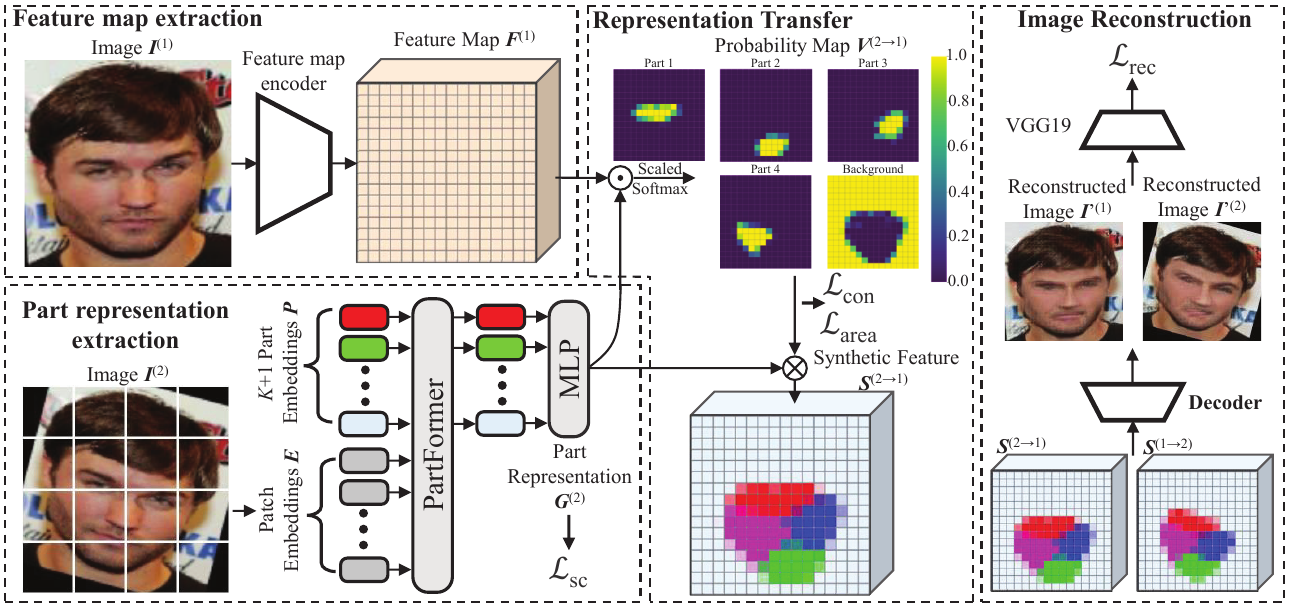}
	\caption{The overall architecture for part-specific attention learning. It takes paired images ($\bm{I}^{(1)}$ and $\bm{I}^{(2)}$), generated from the same image $\bm{I}$ using random geometric transformations, as the inputs. A feature map encoder is utilized to extract dense feature maps $\bm{F}^{\left(n\right)}$, $n=1, 2$; a PartFormer is used to extract the part representations $\bm{G}^{\left(n\right)}$, $n=1, 2$, in a global manner; and a novel transfer module is designed for transferring part representations into dense feature maps. After exchanging the part representations from the paired images, the transfer module first calculates a probability map $\bm{V}^{\left(2 \rightarrow 1\right)}$ based on $\bm{F}^{\left(1\right)}$ and $\bm{G}^{\left(2\right)}$ using Hadamard product, and then normalizes it with a scaled softmax function. Finally, $\bm{G}^{\left(2\right)}$ is assigned to form a synthetic feature map $\bm{S}^{\left(2 \rightarrow 1\right)}$ according to $\bm{V}^{\left(2 \rightarrow 1\right)}$. Similarly, $\bm{S}^{\left(1 \rightarrow 2\right)}$ is generated from $\bm{F}^{\left(2\right)}$ and $\bm{G}^{\left(1\right)}$. Both $\bm{S}^{\left(1 \rightarrow 2\right)}$ and $\bm{S}^{\left(2 \rightarrow 1\right)}$ are fed into a decoder for reconstruction to form a closed loop for part-specific attention learning.}
	\label{fig2}
\end{figure*}

In terms of unsupervised part discovery, the quality of the learned part representations directly determine its performance. Without explicitly learning part representations, Collins et al.~\cite{DEF} utilize matrix factorization to highlight regions with similar semantic within the feature map. As a result, it only predicts rough part masks, without high resolution and accuracy. To explicitly extract part representations, Liu et al.~\cite{LiuDis} extend the disentangling of object shape and appearance from unsupervised landmark detection to unsupervised part discovery. However, they directly use the mean representation of the predicted region as the part representation. However, the learned representation may not adequately describes the corresponding part. To improve the quality of part representations, contrastive learning \cite{NIPSPart}, \cite{Part_ReID}, and clustering \cite{amir2021deep}, are widely used in part discovery to cluster features within the same part and distinguish the features across different parts. As a result, the means of the clusters become better descriptors for their corresponding parts, promising better semantic consistency for the discovered parts. He et al. \cite{GANSeg} further develop GAN for unsupervised part discovery by introducing a set of part embeddings that serve as part representations during generation. Similar to other GAN-based methods, this method also requires two training stages: one for the GAN and another for the segmentation network. Recent studies \cite{Capsule}, \cite{AliaksandrSeg},\cite{PartAssembly} have improved the quality of part representation by considering part pose and have learned part representations by reassembling object parts. However, the training samples should be the consecutive frames or the images under constrained scenarios. Otherwise, these methods tend to identify background regions as object parts during reassembly.

Overall, the quality of the learned part representations directly determines the performance of part discovery. We approach the improvement of part representation quality from a novel perspective by encouraging ViT to learn a part-specific attention mechanism based on the alignment of dual representations. Compared to previous studies, this part-specific attention can be trained end-to-end without the need for consecutive frames or images captured under constrained conditions. It can adaptively generate high-quality part representations. As a result, our method has a broader range of applications while still achieving competitive performance.

\section{Method}

The unsupervised part discovery task aims to train a part detector $\Phi_{\rm part}$ that assigns each image pixel to one of $K$ semantic object parts or one background without any annotations. That is
\begin{equation}
	\Phi_{\rm part}(\bm{I}; \Theta)=\bm{M}_{\rm part} \in \left\{0, 1 \right\}^{H_{\rm I} \times W_{\rm I} \times (K+1)},
\end{equation}
where $\bm{I} \in \mathbb{R}^{H_{\rm I} \times W_{\rm I} \times 3}$ is the input image, ($W_{\rm I}$, $H_{\rm I}$) is the corresponding size of the input image. $\bm{\Theta}$ and $\bm{M}_{\rm part}$ denote network parameters and the predicted mask respectively. Notably, in the setting of unsupervised part discovery, the available data consists solely of the input images $\bm{I}$, without any annotations.

To train such a part discovery model in an unsupervised manner, we design a pretext task based on image reconstruction to learn part representations with good reliability and geometric transformation invariance. We first generate paired images ($\bm{I}^{(1)}$ and $\bm{I}^{(2)}$) from the same input image $\bm{I}$ by randomly translating and rotating $\bm{I}$ with different distances and angles. Then, two subnetworks extract both part representations $\bm{P}$ and dense feature maps $\bm{F}$ from each image. Next, the part representations from two images are exchanged and aligned with the pixel-level semantics contained in the feature maps to reconstruct the corresponding images for unsupervised part representation learning. Finally, during the testing phase, the learned part representations act as a series of detectors, producing high confidence in the corresponding parts on the feature maps for pixel-level part discovery. We detail our method in the rest of this section.

\subsection{Overall Framework}

The overall framework is shown in Fig. 2. It consists of a feature map encoder for pixel representation extraction, a PartFormer for part representation extraction, a novel representation transfer module for dual representations alignment and transferring sparse part representations into a synthetic feature map, and a CNN-based decoder for image reconstruction. The use of the feature map encoder and PartFormer enables our method to decompose input images into two independent features: dense pixel-level features and part-level representations. With the dual representation alignment, these part representations evolve into a series of reliable and adaptive detectors, predicting part discovery masks based on the dense feature maps. This decompose also provides the basis for part representation exchange among paired images, further improving the geometric transformation invariance of learned part representations. Moreover, the synthetic feature map in the representation transfer module serves as a bridge that enables the geometric and semantic constraints to be applied to PartFormer for part-specific attention learning.

\textbf{Feature map encoder}: this encoder aims to produce dense feature maps with pixel-level semantic $\bm{F}^{\left(n\right)} \in \mathbb{R}^{H \times W \times C}$, $n=1, 2$, where $\left(H, W\right)$ is the feature map size and $n$ is the index of the paired images. The functions of the feature maps are twofold. During training, they are fed into the representation transfer module, helping the part representations retain spatial information. With this spatial information, the geometric constraints can be applied to the part representations, and these representations can also be further used in reconstruction to enforce semantic constraints. In the testing phase, the part representations predict the probability map of object parts based on the pixel-level semantics contained in the feature map. In our method, the feature map encoder can be either a CNN or a ViT, initialized from scratch or pretrained on a large-scale dataset. Compared to the datasets used in part discovery, existing large-scale datasets for pretraining, such as ImageNet \cite{ImageNet}, contain various views and object poses. By leveraging the knowledge from these datasets, our method can learn part-specific attention that more closely approximates human perception. As a result, it addresses a shared limitation of previous works: the inability to discern the orientations of nearly symmetric objects.

\textbf{PartFormer}: to learn the specific representations for $K$ part, we propose the PartFormer, replacing the class embedding of a standard ViT by $\left(K+1\right)$ learnable part embeddings $\bm{P} \in \mathbb{R}^{\left(K + 1\right) \times d}$ ($K$ for foreground parts and one for background). $d$ is the number of the dimension in part embeddings. In the PartFormer, the part embeddings aggregate the image patches embeddings $\bm{E}$ adaptively based on the self-attention mechanism to generate the part representations in a global manner. Similar to \cite{DETR}, with the proper constraints, the various weights among part embeddings drive them to specialize on certain object parts. Therefore, the part embeddings also serve as an implicit semantic constraint in this method. Then, the outputs of the PartFormer are fed into a multilayer perceptron (MLP) block for dimensionality reduction. The learned part representations for two input images are $\bm{G}^{\left(n\right)} \in \mathbb{R}^{\left(K + 1\right) \times C}$, $n=1, 2$, where $C$ is the number of the dimension in the feature map output by the feature map encoder.

\textbf{Representation transfer module}: after exchanging the part representations from the paired images, this module transfers the part representations $\bm{G}^{(1)}$ into a synthetic feature map $\bm{S}^{1 \rightarrow 2}$ based on their matching results $\bm{V}^{1 \rightarrow 2}$ with the feature map $\bm{F}^{(2)}$. Similarly, $\bm{S}^{2 \rightarrow 1}$ is generated from $\bm{G}^{(2)}$ and $\bm{V}^{2 \rightarrow 1}$ in the same way. As a result, the synthetic feature map $\bm{S}$ incorporates the part-level features and spatial information of the $\bm{I}^{\left(1\right)}$ and $\bm{I}^{\left(2\right)}$ respectively. This transferring provides the basis for dual representation alignment and part-specific attention learning. By applying geometric to $\bm{V}^{1 \rightarrow 2}$ and $\bm{V}^{2 \rightarrow 1}$, which will be introduced in Section 3.2, the part representations $\bm{G}$ are aligned with the feature map $\bm{F}$. Then, the semantic constraints, which will be introduced in Section 3.3, can be applied to the part representations $\bm{G}$ through the synthetic feature maps $\bm{S}^{1 \rightarrow 2}$ and $\bm{S}^{2 \rightarrow 1}$.

Specifically, after exchanging the part representations $\bm{G}^{\rm (1)}$ and $\bm{G}^{\rm (2)}$ extracted from the paired images, a probability map $\bm{V}^{2 \rightarrow 1}\in\mathbb{R}^{H \times W \times \left(K+1\right)}$ is calculated from $\bm{F}^{\left(1\right)}$ and $\bm{G}^{\left(2\right)}$ as:
\begin{equation}
	V_{i, j, k}^{2 \rightarrow 1} = \frac{e^{\left( {\tau \cdot \bm{G}^{\left(2\right)}_k \bm{F}^{{\left(1\right)}T}_{i, j}  }\right)}}{\sum_{k^\prime=1}^{K+1} e^{\left( \tau \cdot \bm{G}^{\left(2\right)}_{k^\prime} \bm{F}^{{\left(1\right)}T}_{i, j} \right)}},
\end{equation}
where $\bm{F}^{{\left(1\right)}}_{i, j}$ is the feature vector at the position $\left(i, j\right)$ of $\bm{F}^{\left(1\right)}$ and $\bm{G}^{\left(2\right)}_k$ is the representation of the $k$-th part. $\tau$ is a hyperparameter that controls the smoothness of the probability map. $V_{i, j, k}^{2 \rightarrow 1}$ indicates the probability of the $k$-th part at position $\left(i, j\right)$. The geometric constraints can further be applied to $\bm{G}$ through the dense probability map $\bm{V}$ for part specific attention learning. Then, $\bm{G}^{\left(2\right)}$ is assigned form a synthetic dense feature map $\bm{S}^{2 \rightarrow 1}  \in \mathbb{R}^{(H \times W \times C)}$ according to $\bm{V}^{2 \rightarrow 1}$ as follows:
\begin{equation}
	\bm{S}^{2 \rightarrow 1}_{i, j} = \sum_{k=1}^{K+1}\left( V_{i, j, k}^{2 \rightarrow 1} \cdot \bm{G}^{\left(2\right)}_k\right), 
\end{equation}
where $\bm{S}^{2 \rightarrow 1}_{i, j}$ is the feature at position $\left(i, j\right)$ of $\bm{S}^{2 \rightarrow 1}$. Similarly, $\bm{S}^{1 \rightarrow 2}$ is generated from $\bm{F}^{\left(2\right)}$ and $\bm{G}^{\left(1\right)}$.

\textbf{CNN based decoder}: both $\bm{S}^{1 \rightarrow 2}$ and $\bm{S}^{2 \rightarrow 1}$ are fed into a decoder for reconstruction to form a closed loop, which ensures that the part representations to be meaningful and explicitly contain the appearance features of corresponding parts. Additionally, by incorporating part representation exchange, the reconstruction also provides a constraint that encourages the part representations extracted from the paired images to obtain the same appearance features. Therefore, the learned part representations are invariant to geometric transformations. The decoder consists of two upsampling layers, five $3\times 3$ convolution blocks and an additional $3\times 3$ convolutional layer at the end of the decoder. Since $\bm{G}_k$ is normalized with LayerNorm (LN) layer \cite{LN} in the ViT, we replace the BatchNorm (BN) layer \cite{BN} in each convolution block with a LN layer, as in previous work \cite{2020S}, to maintain consistency in normalization for better reconstruction results. Therefore, each convolution block contains a convolution layer, followed by a ReLU layer \cite{ReLU} and a LN layer.

\subsection{Geometric Constraints}

In this paper, we utilize two loss functions: concentration loss and area loss to provide geometric constraints.

\textbf{Concentration loss}, on the one hand, aligns the part representations with the pixel representations. Thus, the part representations exhibit higher similarity with the pixel representations in the corresponding part regions and lower similarity with those in other regions. On the other hand, it encourages \textit{foreground} pixels with the same semantics on the probability map $\bm{V}$ to form a concentrated and connected part, which drives the PartFormer to focus locally and to form part-specific attention. Otherwise, the global receptive field of the PartFormer might cause the model to degrade into a color cluster model. The concentration loss $\mathcal{L}_{\rm con}$ can be written as:

\begin{equation}
	\mathcal{L}_{\rm con} = \sum_{k=1}^{K}\sum_{i=1, j=1}^{W, H}\left\lVert \begin{bmatrix}i \\ j\end{bmatrix}  - \begin{bmatrix}c_{\rm I}^k \\ c_{\rm J}^k\end{bmatrix} \right\rVert_2^2 V_{i,j,k} / z_k,
\end{equation}
where $\left(c_{\rm I}^k, c_{\rm J}^k\right)$ is the center of the $k$-th part. $z_k = \sum_{i=1, j=1}^{W, H} V_{i,j,k} + \epsilon$ is the predicted area size for $k$-th part, and $\epsilon$ is a small constant to prevent $z_k$ to be zero.

%$z_k$ is the predicted area size for $k$-th part, serving as a normalization factor in eq. 5. $\left(c_{\rm I}^k, c_{\rm J}^k\right)$ and $z_k$ can be computed as:
\begin{equation}
	\begin{bmatrix}c_{\rm I}^k \\ c_{\rm J}^k\end{bmatrix} = \sum_{i=1, j=1}^{W, H} \begin{bmatrix}i \\ j\end{bmatrix} V_{i,j,k} / z_k,
\end{equation}

%To satisfy this loss, the part 

% 

\textbf{Area loss} limits the minimum area size of  of \textit{foreground} parts and \textit{background} on $\bm{V}$. Otherwise, the training process will prioritizes the concentration constraint by describing the entire image using the representation for background. As a result, the training result gets into a trivial solution. The novel area loss $\mathcal{L}_{\rm area}$ can be written as:

\begin{equation}
	\mathcal{L}_{\rm area} = \sum_{k=1}^{K+1}\frac{1}{1+z_k/ \alpha},
\end{equation}
where $\alpha$ is a hyperparameter. When the predicted area size $z_k$ is less than a certain threshold, $\mathcal{L}_{\rm area}$ will increase dramatically. Therefore, the predicted area size is expected to be larger than this threshold, which prevents the model from getting into trivial solutions. This threshold is determined by $\alpha$. Therefore, $\alpha$ serves as prior knowledge, determining the expected minimum size of each predicted area.

\subsection{Semantic Constraints}

Semantic constraints are another kind of essential constraint to parts discovery. The semantic constraints include the reconstruction loss and the semantic consistency loss.

\textbf{Reconstruction loss} penalizes the difference between the input images ($\bm{I}^{\left(1\right)}$ and $\bm{I}^{\left(2\right)}$) and the reconstructed images ($\bm{I}^{\prime\left(1\right)}$ and $\bm{I}^{\prime\left(2\right)}$). We use a pretrained VGG19 $\Phi$ \cite{VGG} with frozen weights to extract the intermediate features from the input and reconstructed images and measure their semantic difference by perceptual loss \cite{Perceptual}. The difference $\mathcal{L}_{\rm rec}$ can be calculated as:
\begin{equation}
	\mathcal{L}_{\rm rec} = \Vert \Phi(\bm{I}^{(1)}) - \Phi(\bm{I}^{\prime(1)})\ \Vert_{1} + \Vert \Phi(\bm{I}^{(2)}) - \Phi(\bm{I}^{\prime(2)})\ \Vert_{1}.
\end{equation}
Because of the dual representation alignment, the information of each part region in $\bm{S}^{1 \rightarrow 2}$ and $\bm{S}^{2 \rightarrow 1}$ is mainly from its corresponding part representation. Therefore, this constraint encourages the information of each part region in $\bm{I}^{(1)}$ and $\bm{I}^{(2)}$ to be explicitly contained in its corresponding part representations. Moreover, the reconstruction incorporated with representation exchange also encourages the part representations extracted from $\bm{I}^{(1)}$ and $\bm{I}^{(2)}$ to be consistent, which guarantees better geometric transformation invariance for the learned part representations.

Some works \cite{TUSK} also use the mean squared error (MSE) to measure the difference between the input images and reconstructed images. However, minimizing this difference only drives our method to recover the color of specific regions, rather than extracting part representations with corresponding part information. Consequently, the model trained with MSE loss fails to discover the object parts with specific semantics.

\textbf{Semantic consistency loss} aims to encourage the representations of the same part to get close in the latent space, while pushing the representations of different parts to be orthonormal on the hypersphere. Inspired by Liu et al. \cite{LiuDis}, we utilize ArcFace loss \cite{ArcFace} to reinforce semantic consistency. It defines $K$ learnable vectors with dimension $C$ for $K$ \textit{foreground} parts, which can be written as $\bm{W} \in \mathbb{R}^{K \times C}$. $\bm{W}$ is shared across all samples. The angle between the $t$-th vector of $\bm{W}$ and the $k$-th representation of $\bm{G}$ is defined as $\theta_{(t, k)}$. Then, the semantic consistency loss $\mathcal{L}_{\rm sc}$ is computed as:
\begin{equation}
	\mathcal{L}_{\rm sc} = -\frac{1}{K}\sum_{k=1}^{K}\log\frac{e^{s\left(\cos\left(\theta_{\left(k, k\right)}+m\right)\right)}}{e^{s\left(\cos\left(\theta_{\left(k, k\right)}+m\right)\right)} + \sum_{t=1, t\neq k}^{K}e^{s\cos\theta_{\left(t, k\right)}}},
\end{equation}
where $m$ and $s$ are two hyperparameters for penalizing angular margin and scaling radius. $m$ ensures that the loss function still has a certain gradient when $\theta_{(k, k)}$ is very small, and $s$ determines the smoothness of the semantic consistency loss. On one hand, the semantic consistency loss encourages each part representation of all training samples to approximate its corresponding learnable vector in cosine distance by minimizing the $\theta_{(k, k)}$. As a result, the representations of the same part extracted from different images obtain more consistent semantics. On the other hand, the semantic consistency loss maximizes the $\theta_{(t, k), (t \neq k)}$ to encourage the representations of different parts to be orthonormal. This allows the part representations of different parts to be more easily distinguished in the latent space.

Finally, the overall training object can be written as:
\begin{equation}
	\mathcal{L} = \mathcal{L}_{\rm rec} + \lambda_{\rm sc}\mathcal{L}_{\rm sc} + \lambda_{\rm con}\mathcal{L}_{\rm con} + \lambda_{\rm area}\mathcal{L}_{\rm area},
\end{equation}
where $\lambda_{\rm sc}$, $\lambda_{\rm con}$ and $\lambda_{\rm area}$ are the weights of $\mathcal{L}_{\rm sc}$, $\mathcal{L}_{\rm con}$ and $\mathcal{L}_{\rm area}$ respectively.

\begin{figure}[t!]
	\centering
	\includegraphics[width=\linewidth]{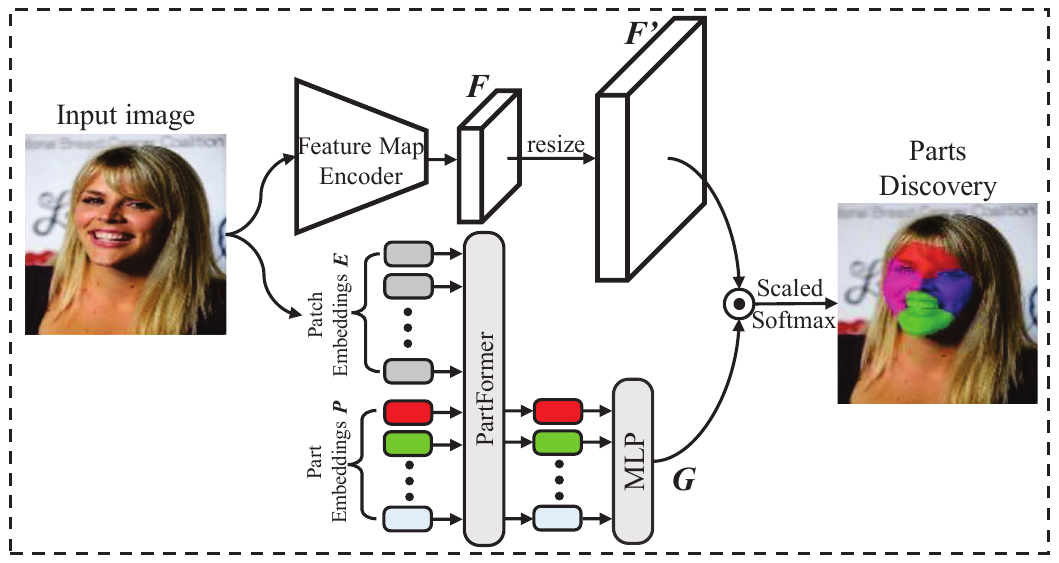}
	\caption{Overall pipeline for parts discovery in testing phase. The weights of the feature map encoder and PartFormer come from the two-stream architecture.}
	\label{fig3}
\end{figure}

\subsection{Part Discovery}
We further predict pixel-level part masks using the feature map encoder and PartFormer, both trained within the proposed two-stream architecture, as shown in Fig. 3. Because the part representations and pixel representations have been aligned during training, the part representations can serve as a series of detectors, outputting high confidence for the pixels with the similar semantics. The learned part-specific attention enables the part representations to adaptively describe the corresponding target parts on various conditions, which further improves the robustness of part discovery. Specifically, the feature map encoder first extracts a feature map with pixel-level semantics $\bm{F}$, while the PartFormer extracts the part representations $\bm{G}$ from the same input image $\bm{I} \in \mathbb{R}^{H_{\rm I} \times W_{\rm I} \times 3}$. Then, $\bm{F}$ is resized from the original size $\left(H, W\right)$ to the image size $\left(H_{\rm I}, W_{\rm I}\right)$ by bilinear interpolation. The interpolation allows the proposed method to produce part masks at a higher resolution. As a result, the discovered parts typically exhibit smoother and tighter boundaries. Finally, the probability map is calculated from the resized shape feature $\bm{F}^\prime$ and part representations $\bm{G}$, as described in Eq. (2), and serves as the predicted part masks.

\section{Experiments}
\subsection{Datasets}
Experiments are carried out on five widely used datasets to verify the effectiveness of the proposed method.

\textbf{CelebA-in-the-wild} \cite{CelebA} consists of 162,770 samples for training, 19,867 for validation and 19,962 for testing. Each sample is under unconstrained conditions and annotated with five landmarks. Following \cite{SCOPS}, we filter out the \textit{unaligned} images where the face covers less than 30\% of the pixel area from the training set. Additionally, the samples from MAFL \cite{MAFL} (a subset of CelebA) test set are also excluded as \cite{ZhangLandmark}, resulting in 45,609 samples for training. Then, this setting employs the train set of MAFL (5,397 samples) for validation and the test set of MAFL (283 samples) for testing.

\textbf{AFLW} \cite{AFLW} comprises 25,993 faces captured in the wild, each with annotated with 21 landmarks manually. These faces exhibit a large variety in appearance, such as gender, age, pose, and expression. Thewlis et al. \cite{AFLW2} further propose a subset of AFLW, which contains 10,122 \textit{unaligned} samples for training and 2,991 \textit{unaligned} samples for test. Each sample in this subset is annotated with 5 landmarks.

\textbf{CUB-2011} \cite{CUB} is a bird dataset that comprises 5,994 samples for training and 5,794 samples for test. Each sample is annotated with 15 landmarks, the bounding box and the species label. We also calculate a transformation matrix, using the predicted mask centers to map to the 15 annotated landmarks on the training set.

\textbf{DeepFashion} \cite{Deepfashion} is a large-scale fashion dataset, including more than 800,000 images. These images consist of well-posed shop images and unconstrained consumer photos. Each sample in the ``In-shop Clothes Retrieval Benchmark" subset of the DeepFashion is annotated with a parsing mask and dense pose. Following \cite{NIPSPart}, we utilize the ``In-shop Clothes Retrieval Benchmark" from Deepfashion for training and testing.

\textbf{PartImageNet} \cite{PartImageNet} consists 158 object classes with large variations in appearance, pose and scenario. These categories belong to 11 different super-categories, such as car, quadruped and biped. We follow the setting in \cite{PDiscoNet}, using 109 classes, which includes 14,876 images for training and 1,664 images for testing, to validate the effectiveness of the proposed method. Each image is manually annotated with the corresponding part segmentation maps.

\begin{table*}[t!]
	\centering
	\begin{tabular}{|m{2.2cm}<{\centering}|m{1.2cm}<{\centering}|m{1.4cm}<{\centering}|m{1.0cm}<{\centering}|m{1.0cm}<{\centering}|m{1.0cm}<{\centering}|m{1.0cm}<{\centering}|m{1.0cm}<{\centering}|m{1.0cm}<{\centering}|}
		\hline
		\multirow{2}{*}{Method} & \multirow{2}{*}{Type} & Auxiliary & \multicolumn{2}{c|}{Inter-ocular NME $\downarrow$} & \multicolumn{2}{c|}{NMI $\uparrow$} & \multicolumn{2}{c|}{ARI $\uparrow$}\\ \cline{4-9}
		&  & Supervision & K=4 & K=8 & K=4 & K=8 & K=4 & K=8\\ \hline
		Thewlis et al. \cite{ULD}& landmark  & N/A  & - &  31.30\% & - & -& -& - \\
		Zhang et al. \cite{ZhangLandmark} & landmark & N/A & - & 40.83\% & - & -& -& - \\
		Lorenz et al. \cite{LOR_LANDMARK} & landmark & N/A & 15.49\% & 11.41\% & - & -& -& - \\
		IMM \cite{JakabLandmark} & landmark & N/A & 19.42\% & 8.74\% & - & -& -& -  \\ \hline
		DFF \cite{DEF} & part  & N/A & - & 31.30\% & - & -& -& - \\
		SCOPS \cite{SCOPS} & part & N/A & 46.62\% & 22.11\% & - & -& -& - \\
		SCOPS \cite{SCOPS} & part & saliency & 21.76\% &15.01\% & - & -& -& - \\
		Liu et al. \cite{LiuDis} & part & N/A & 15.39\%  & 12.26\% & - & -& -& - \\
		Huang et al \cite{HuangAttention} & part & attribute & {\color{red} \textbf{8.75\%}} & {\color{blue} \textbf{7.96\%}} & {\color{blue} \textbf{56.69}} & 54.80 & {\color{blue} \textbf{34.74}} & 34.74 \\
		GANSeg \cite{GANSeg} & part & N/A & 12.26\% & {\color{red} \textbf{6.18\%}} & 41.71 & {\color{red} \textbf{67.28}} & 28.06 & {\color{red} \textbf{56.23}}\\ 
		DINO \cite{amir2021deep} & part & N/A & 11.36\% & 10.74\% & 1.38 & 1.12 & 0.01 & 0.01 \\
		PDiscoNet \cite{PDiscoNet} & part & identity & {\color{blue} \textbf{11.11\%}} & 9.82\% & {\color{red} \textbf{75.97}} & 62.61 & {\color{red} \textbf{69.53}} & 51.89 \\
		\hline
		Ours & part & N/A & 13.28\% & 8.70\%  & 37.12 & {\color{blue} \textbf{62.91}} & 25.42 & {\color{blue} \textbf{55.82}} \\ 
		Ours$^\star$ & part & N/A & 12.28\% & 10.24\% & 50.69 & 58.35 & 34.57 & 46.78 \\ \hline
	\end{tabular}
	\caption{Inter-ocular NME of different methods with $K=4$ and $K=8$ on CelebA-in-the-wild. Key: [{\color{red} \textbf{Best}}, {\color{blue} \textbf{Second best}}, $^\star$=pretrained on ImageNet using DINO]}
	\label{Table1}
\end{table*}

\subsection{Evaluation Metrics}

\begin{figure}[t!]
	\centering
	\includegraphics[width=\linewidth]{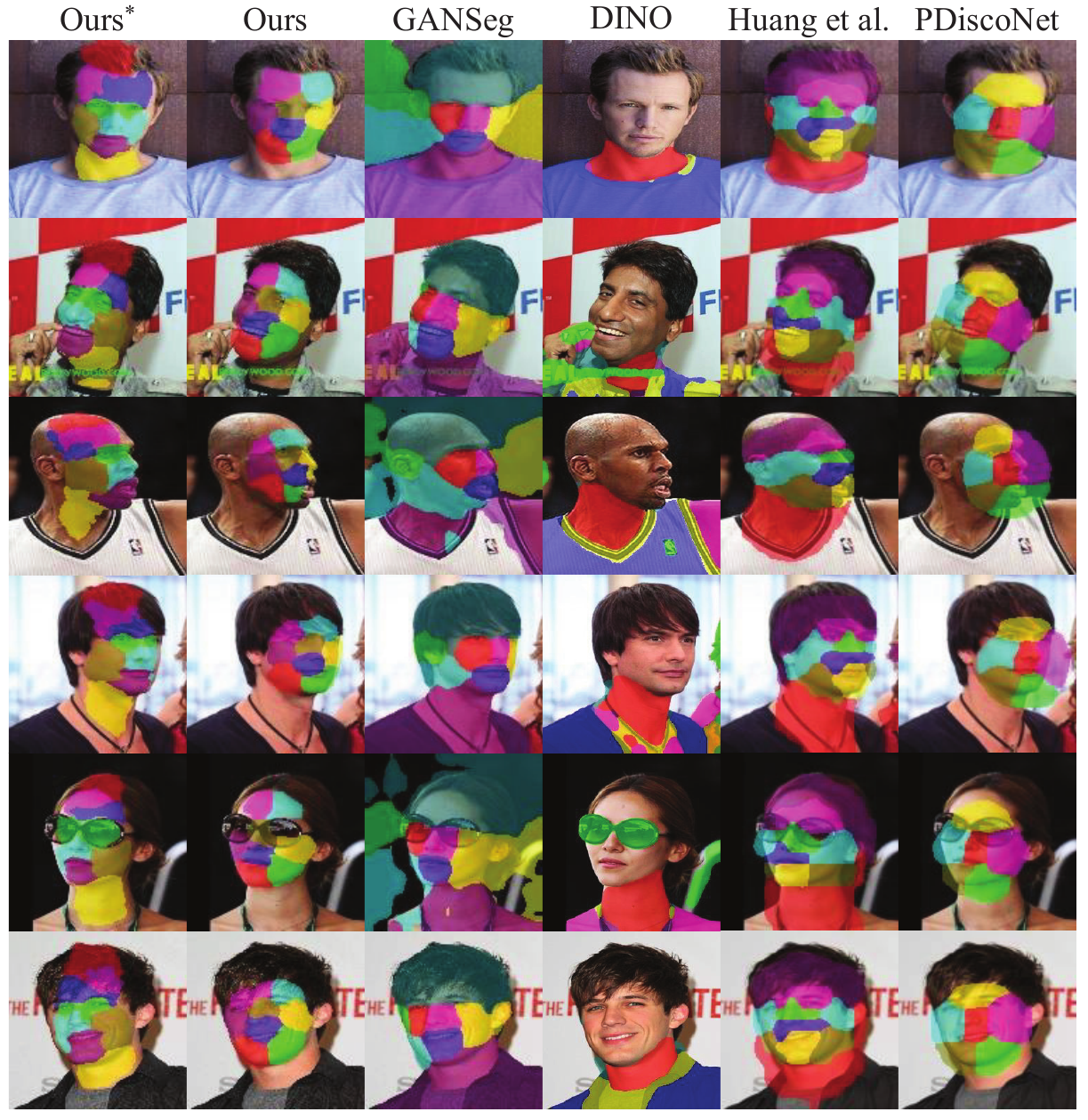}
	\caption{Visualized part discovery results of our proposed method and other state-of-the-art methods on CelebA-in-the-wild (K=8). Key: [\textcolor[rgb]{1, 0, 0}{\textbf{Part1}}, \textcolor[rgb]{0, 1, 0}{\textbf{Part 2}}, \textcolor[rgb]{0, 0, 1}{\textbf{Part 3}}, \textcolor[rgb]{1, 0, 1}{\textbf{Part 4}}, \textcolor[rgb]{0, 1, 1}{\textbf{Part 5}}, \textcolor[rgb]{1, 1, 0}{\textbf{Part 6}}, \textcolor[rgb]{0.5, 0.5, 0}{\textbf{Part 7}}, \textcolor[rgb]{0.5, 0.0, 0.5}{\textbf{Part 8}}]}
	\label{fig4}
\end{figure}

\textbf{Normalized Mean Error (NME)}: CelebA-in-the-wild, AFLW, and CUB are labeled with keypoints, which can be viewed as a form of part annotation. To evaluate the semantic consistency of the discovered parts, a transformation matrix can be calculated on the training set or validation set to project the centroids of the discovered parts onto the labeled landmarks. Then, the normalized mean error between the transformed part centroids and labeled keypoints on the test set can be utilized to measure the semantic consistency. The NME is calculated as follows:
\begin{equation}
	NME = \frac{1}{K^{\rm gt}}\sum_{k^\prime=1}^{K^{\rm gt}}\left\lVert\frac{\bm{C}_{\rm T}^{k^\prime} - \bm{P}_{\rm gt}^{k^\prime}}{d_{\rm norm}} \right\lVert,
\end{equation}
where $K^{\rm gt}$ is the number of labeled keypoints. $\bm{C}_{\rm T}^{k^\prime}$ and $\bm{P}_{\rm gt}^{k^\prime}$ are the $k^\prime$-th transformed part centroid and labeled landmark respectively. $d_{\rm norm}$ is the normalization factor. We utilize the inter-ocular distance as $d_{\rm norm}$ for CelebA-in-the-wild and AFLW, and the annotated bounding box as $d_{\rm norm}$ for CUB-2011.

\textbf{Normalized Mutual information (NMI) and Adjust Rand Index (ARI)}: NMI and ARI are two commonly used metrics that reflect the similarity between two clusterings, even when the number of clusters differs. Therefore, they are utilized to measure the semantic consistency between results predicted by unsupervised models and annotations in unsupervised part discovery. They are calculated based on manually labeled keypoints on CelebA-in-the-wild. AFLW and CUB, as well as annotated masks on Deepfashion and PartImageNet.

\subsection{Implementation Details}
We implement the proposed methods both with and without the pretrained weight. For the model without pretraining, we utilize the stages before \textit{layer4} of a trainable ResNet18 \cite{Resnet} to extract the feature map with pixel-level semantics. For the model with pretraining, we employ a \textit{frozen} ViT-S/8 \cite{VIT} pretrained with DINO \cite{DINO} as the feature map encoder. This frozen ViT-S/8 is further followed by three trainable CNN blocks for dimensionality reduction. Each CNN block consists of a $3\times 3$ CNN layer, a ReLU layer \cite{ReLU} and a Batchnorm layer \cite{BN}. For the PartFormer, we set the number of layers to 6, the number of heads to 8, hidden size to 256, MLP size to 1024 and patch size to $16 \times 16$. The PartFormer is initialized from scratch for both models, with and without pretraining. The input size is $128 \times 128$ for the model without pretraining and $256 \times 256$ for the model with pretraining, resulting in that the size of $\bm{V}$ and $\bm{S}$ is $32 \times 32$ for both implementations.

\begin{figure*}[t!]
	\centering
	\includegraphics[width=0.92\linewidth]{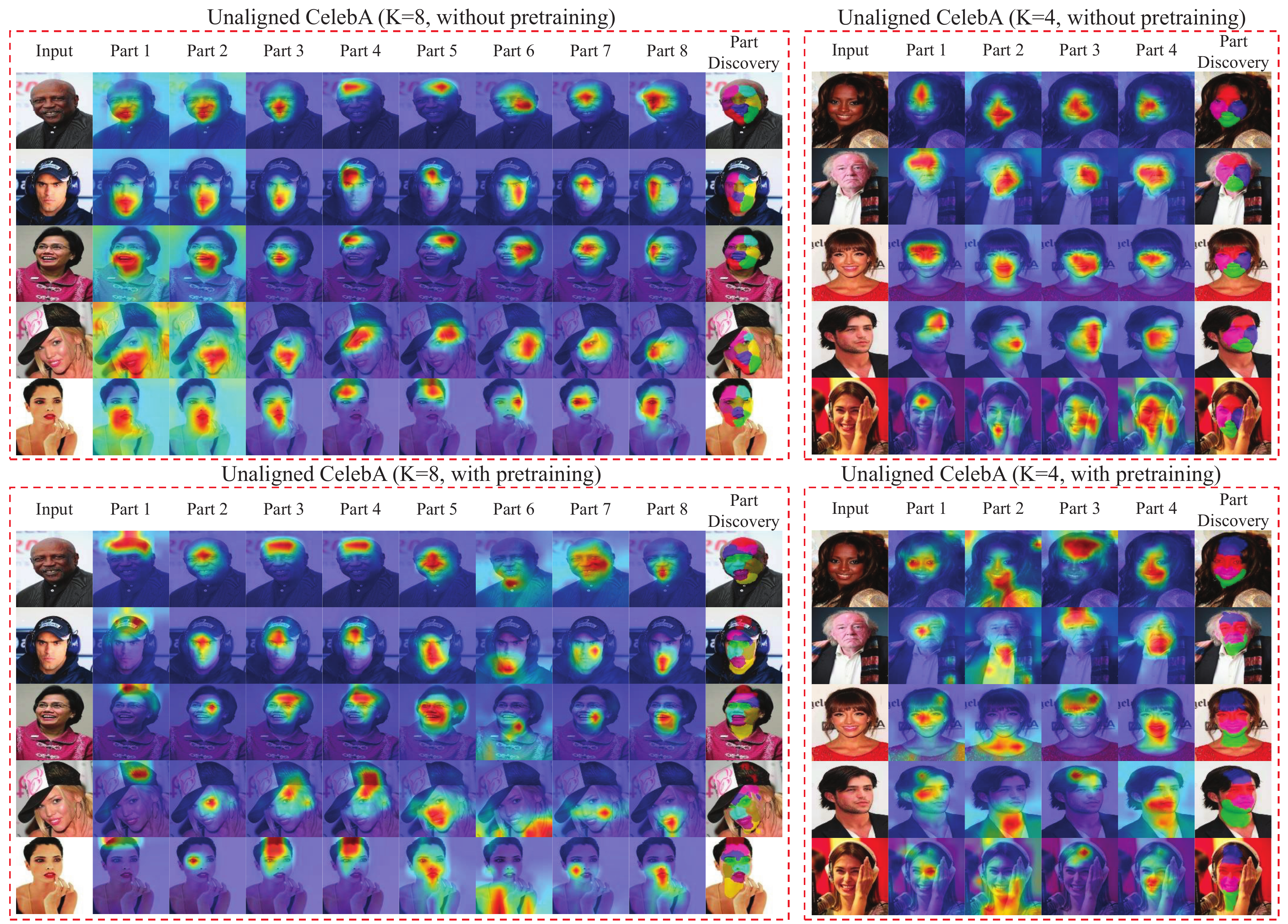}
	\caption{Results of the parts discovery and the corresponding attention maps for different discovered parts on CelebA-in-the-wild (K=4/8, with/without pretraining). Key: [\textcolor[rgb]{1, 0, 0}{\textbf{Part1}}, \textcolor[rgb]{0, 1, 0}{\textbf{Part 2}}, \textcolor[rgb]{0, 0, 1}{\textbf{Part 3}}, \textcolor[rgb]{1, 0, 1}{\textbf{Part 4}}, \textcolor[rgb]{0, 1, 1}{\textbf{Part 5}}, \textcolor[rgb]{1, 1, 0}{\textbf{Part 6}}, \textcolor[rgb]{0.5, 0.5, 0}{\textbf{Part 7}}, \textcolor[rgb]{0.5, 0.0, 0.5}{\textbf{Part 8}}]}
	\label{fig5}
\end{figure*}

To generate paired images using geometric transformation, we randomly scale ($\pm 5\%$), rotate ($\pm 15^\circ$) and translate ($\pm 5px$) each input image twice during the training. The two-stream architecture is trained by AdamW \cite{AdamW} optimizer with a learning rate of $5\times10^{-4}$. Moreover, the batch size is set to 32. For the model \textit{with} pretraining, we set $\lambda_{\rm con}$ to 0.3, $\lambda_{\rm area}$ to 0.5, $\lambda_{\rm sc}$ to 0.01, and $\tau$ is set to 0.8 (The influences of different values of $\tau$, $\lambda_{\rm sc}$, $K$, and the number of PartFormer layers are discussed in the supplementary file). $s$ and $m$ are set to 20 and 0.5 respectively, following the setting in \cite{LiuDis}. For the model \textit{without} pretraining, it does not contain any prior knowledge about object parts. Therefore, we have to employ a relatively large $\lambda_{\rm con}$, setting $\lambda_{\rm con}$ to 0.5, to encourage it to form part-level attention. Other hyperparameters keep same as the hyperparameters used in the model with pretraining.

\begin{table}[t!]
	\centering
	{\resizebox{\linewidth}{!}{
			\begin{tabular}{|m{2.1cm}<{\centering}|m{1.2cm}<{\centering}|m{1.4cm}<{\centering}|m{0.9cm}<{\centering}|m{0.9cm}<{\centering}|m{0.9cm}<{\centering}|}
				\hline
				\multirow{2}{*}{Method} & \multirow{2}{*}{type} & Auxiliary & \multicolumn{3}{c|}{K=8} \\ \cline{4-6}
				& & Supervision & NME $\downarrow$ & NMI $\uparrow$ & ARI $\uparrow$ \\ \hline
				IMM \cite{JakabLandmark} & landmark & N/A & 13.31\% & - & - \\
				Lorenz et al. \cite{LOR_LANDMARK} & landmark & N/A & 13.60\% & - & - \\ \hline
				SCOPS \cite{SCOPS} & part & N/A  & 56.08\% & - & - \\
				SCOPS \cite{SCOPS} & part & saliency  & 16.05\%  & - & - \\
				Liu et al. \cite{LiuDis} & part & N/A & 13.13\% & - & - \\ 
				GANSeg \cite{GANSeg} & Part & N/A & 11.16\% & 60.77 & 51.94 \\ \hline
				Ours$^\dag$ & part & N/A & 11.87\% & 37.47 & 22.09 \\ 
				Ours & part & N/A & 11.83\% & 54.40 & 40.99 \\ 
				Ours$^\star$ $^\dag$  & part & N/A & {\color{blue} \textbf{11.13}}\% & {\color{blue} \textbf{62.18}} & {\color{blue} \textbf{53.50}} \\ 
				Ours$^\star$ & part & N/A & {\color{red} \textbf{9.89}}\% & {\color{red} \textbf{67.17}} & {\color{red} \textbf{55.31}} \\ \hline
			\end{tabular}
	}}
	\caption{Inter-ocular NME of different methods with $K=8$ on AFLW test set. Key: [$^\dag$=trained on CelebA-in-the-wild, $^\star$=pretrained on ImageNet using DINO]}
	\label{Table2}
\end{table}

\subsection{Comparison with State-of-the-Art Methods}

\textbf{CelebA-in-the-wild}: we tabulate the NME of the proposed method, both with and without pretraining, and compare it with other state-of-the-art methods on CelebA-in-the-wild in Table 1. The corresponding visualized results are shown in Fig. 4. Additionally, we randomly select some images and demonstrate the corresponding part discovery results and attention maps predicted by our method in Fig. 5. On this benchmark, the proposed model without pretraining achieves better performance in the setting of $K=8$. The primary reason for this is that the semantics of the face region learned by the frozen DINO are highly similar, leading to unstable predicted part boundaries. Consequently, in the setting of $K=8$, the pretrained model outperforms the non-pretrained model by 7.81\% and 19.32\% in NMI and ARI respectively. Moreover, because of the high similarity of face features, the model with pretraining cannot divide face region into target number of parts and identifies other regions, such as the hair and neck, as foreground parts. The larger deformation of the hair and neck leads to an unstable part center, which further results in performance degradation in target landmark regression, as measured using NME. Despite failing to outperform the model without pretraining, the visualized results (\textbf{see 3-rd and 5-th rows in Fig. 4}) show that the model with pretraining achieves better robustness in cases of profile views and occlusions.

\begin{table*}[t!]{
		\centering
		\resizebox{1.00\textwidth}{!}{
			\begin{tabular}{|m{2.7cm}<{\centering}|m{0.9cm}<{\centering}|m{1.4cm}<{\centering}|m{0.9cm}<{\centering}|m{0.9cm}<{\centering}|m{0.9cm}<{\centering}|m{0.9cm}<{\centering}|m{1.1cm}<{\centering}|m{1.1cm}<{\centering}|m{1.1cm}<{\centering}|m{1.1cm}<{\centering}|}
				\hline
				\multirow{2}{*}{Method} & \multirow{2}{*}{Pretrain} & Auxiliary  & \multicolumn{4}{c|}{NME (bounding box) $\downarrow$} & \multirow{2}{*}{FG-NMI$\uparrow$} & \multirow{2}{*}{FG-ARI$\uparrow$} & \multirow{2}{*}{NMI$\uparrow$} & \multirow{2}{*}{ARI$\uparrow$} \\ \cline{4-7}
				&  & Supervision & CUB-1 & CUB-2 & CUB-3 & CUB-F & & & & \\ \hline
				SCOPS \cite{SCOPS} & N & saliency & 18.5\% & 18.8\% & 21.1\% & - & - & - & - & - \\
				SCOPS \cite{SCOPS} & N & foreground & 18.3\% & 17.7\% & 17.0\% & 12.6\% & 39.1 & 17.9 & 24.4 & 7.1 \\
				Huang et al \cite{HuangAttention} & Y & species & 15.1\% & 17.1\% & 15.7\% & 11.6\% & - & - & 29.7 & 14.0 \\
				Choudhury et al. \cite{NIPSPart} & Y & foreground & {\color{blue} \textbf{11.3\%}} & {\color{blue} \textbf{15.0\%}} & {\color{red} \textbf{10.6\%}} & {\color{blue} \textbf{9.2\%}} & {\color{red} \textbf{46.0}} & {\color{red} \textbf{21.0}} & {\color{blue} \textbf{43.5}} & {\color{red} \textbf{19.6}} \\ 
				PDiscoNet \cite{PDiscoNet} & Y & species & - & - & - & {\color{red} \textbf{9.1\%}} & - & - & 37.8 & 15.3 \\  \hline
				ULD \cite{ULD, ZhangLandmark} & N & N/A  & 30.1\% &  29.4\% & 28.2\% & - & - & - & - & - \\
				DFF \cite{DEF} & Y & N/A & 22.4\% & 21.6\% & 22.0\% & - & 32.4 & 14.3 & 25.9 & 12.4 \\
				Liu et al. \cite{LiuDis} & N & N/A & 18.2\% & 17.5\% & 19.4\% & - & - & - & - & - \\ 
				GANSeg \cite{GANSeg} & N & N/A & 22.1\% & 22.3\% & 21.5\% & 12.1\% & - & - & - & - \\
				DINO et al. \cite{amir2021deep} & Y & N/A & 17.1\% & 14.7\% & 19.6\% & - & 39.4 & 19.2 & 38.9 & 16.1 \\  \hline
				Ours & N & N/A & 17.0\% & 16.5\% & 20.4\% & 16.7\% & 22.0 & 10.5 & 22.0 & 9.3 \\ 
				Ours$^\star$ & Y & N/A & {\color{red} \textbf{9.9\%}} & {\color{red} \textbf{12.8\%}} & {\color{blue} \textbf{10.7\%}} & 9.9\% & {\color{blue} \textbf{44.8}} & {\color{blue} \textbf{19.3}} & {\color{red} \textbf{45.2}} & {\color{blue} \textbf{19.4}} \\ \hline
			\end{tabular}
	}}
	\caption{Comparisons of the proposed models, both with and without pretraining, on CUB-1/2/3 and CUB-F ($K=4$). The landmark regression error NME, as well as (FG-)NMI and (FG-)ARI metrics are reported. Key: [{\color{red} \textbf{Best}}, {\color{blue} \textbf{Second best}}, $^\star$=pretrained on ImageNet using DINO]}
	\label{Table3}
\end{table*}

Despite using the same backbone as our methods, directly predicting foreground masks using attention map of DINO and clustering the features for part discovery as \cite{amir2021deep} does not work well on CelebA. The main reason is that the attention of DINO views the face region as background and picks up clothes as the foreground region, as shown in Fig. 4. Notably, even though \cite{amir2021deep} fails to decompose face into several parts, it still outperforms many other methods in the metric of NME, while it achieves only 1.38/1.12 and 0.01/0.01 on NMI and ARI respectively, in the settings of $K=4/8$. This result further supports the conclusion in \cite{NIPSPart} that NMI and ARI can better measure the performance of unsupervised part discovery than NME.

\begin{figure}[t!]
	\centering
	\includegraphics[width=\linewidth]{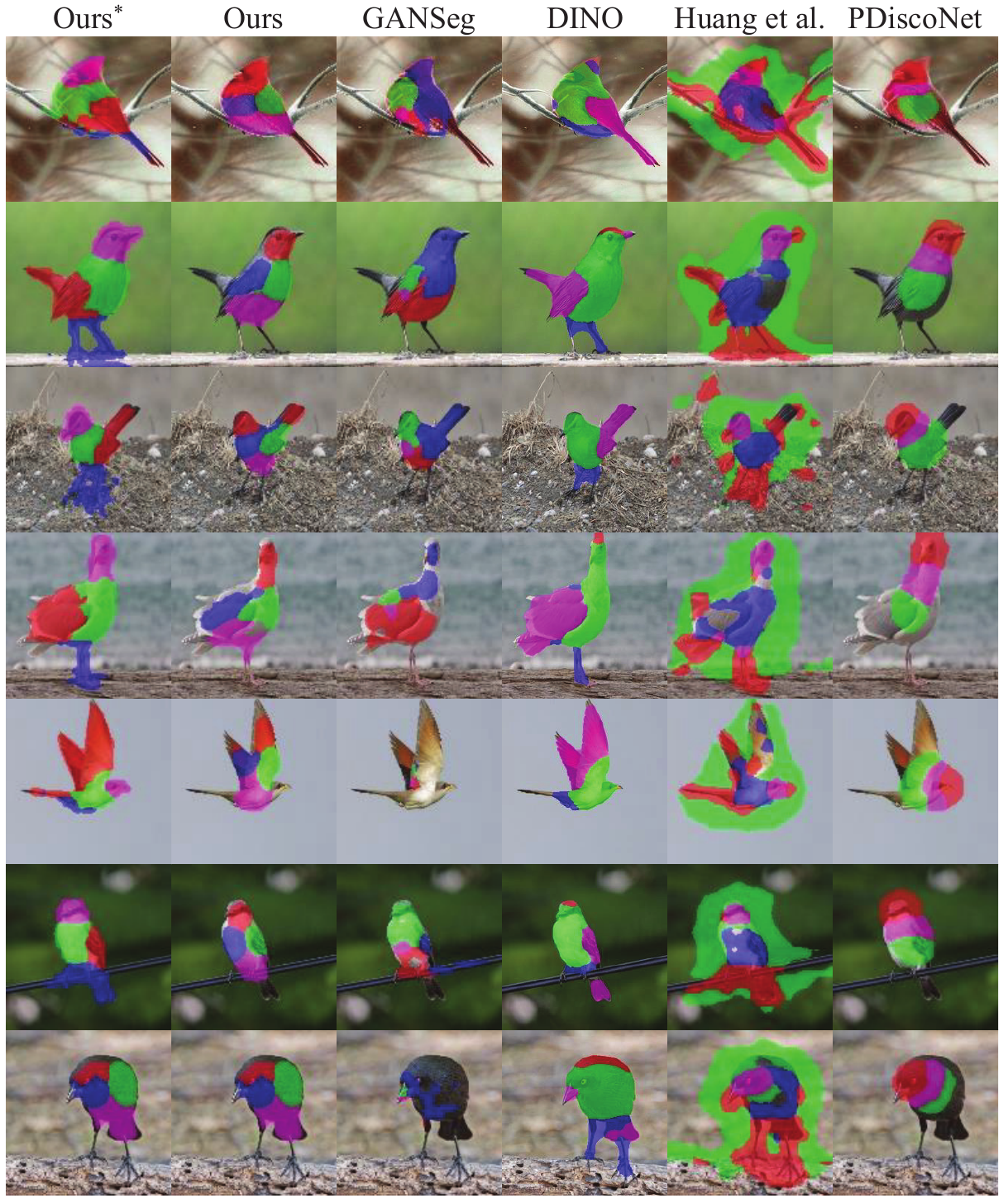}
	\caption{Part discovery results on CUB-F predicted by our method and other state-of-the-art methods. Key: [\textcolor[rgb]{1, 0, 0}{\textbf{Part1}}, \textcolor[rgb]{0, 1, 0}{\textbf{Part 2}}, \textcolor[rgb]{0, 0, 1}{\textbf{Part 3}}, \textcolor[rgb]{1, 0, 1}{\textbf{Part 4}}, $^\star$=pretrained on ImageNet using DINO]}
	\label{fig6}
\end{figure}

Unlike other methods trained end-to-end, the training process of GANSeg consists of two stages. It first generates pseudo labels using a GAN, and then trains a segmentation model with these pseudo labels. The second stage can filter out the noise present in the pseudo labels, leading to more stable prediction results on easy samples. Nevertheless, when it comes to challenging samples (\textbf{see 3-rd and 5-th rows in Fig. 4}), GANseg still fails to identify face regions from the background due to the domain gap between synthetic and real faces. With extra labels for weakly-supervised part discovery, Huang et al. and PDiscoNet respond more effectively to specific regions on faces, resulting in much higher performance in the setting of $K=4$. Nevertheless, the masks predicted by Huang et al. and PDiscoNet only contain rough regions without explicit boundaries for the corresponding face parts (\textbf{see 5-th and 6-th columns in Fig. 4}). Since CelebA only provides five landmarks rather than dense masks for NMI and ARI calculation, this limitation cannot be well reflected in these two metrics.

\begin{figure*}[t!]
	\centering
	\includegraphics[width=\linewidth]{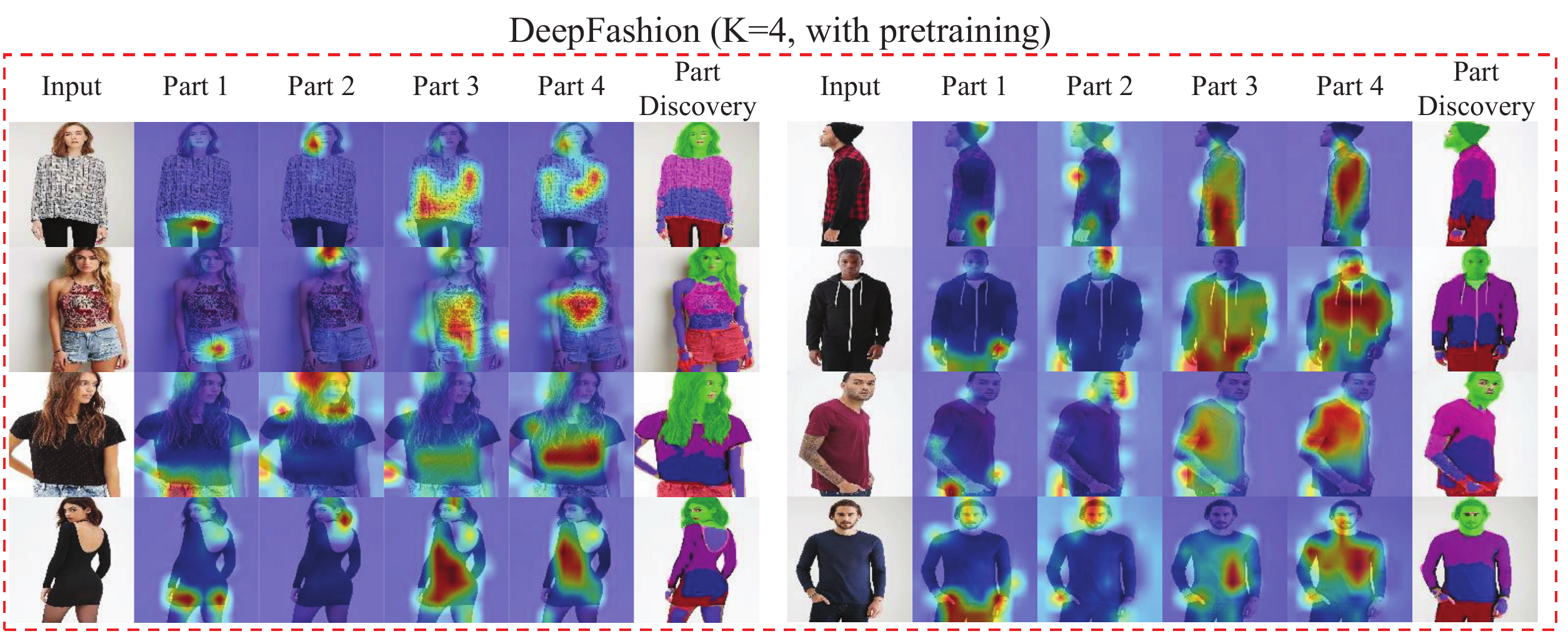}
	\caption{Results of the parts discovery and the corresponding attention maps for different discovered parts on DeepFashion (K=4). Key: [\textcolor[rgb]{1, 0, 0}{\textbf{Part1}}, \textcolor[rgb]{0, 1, 0}{\textbf{Part 2}}, \textcolor[rgb]{0, 0, 1}{\textbf{Part 3}}, \textcolor[rgb]{1, 0, 1}{\textbf{Part 4}}]}
	\label{fig7}
\end{figure*}

\textbf{AFLW}: compared to CelebA, AFLW is more challenging because its samples are typically with heavy occlusion, illumination variance and profile view. We carry out two experiments on AFLW dataset. The first one is \textit{within}-dataset validation. We train the proposed model and calculate the transformation matrix on the AFLW training set. Then, we measure the performance on the test set using NME, NMI and ARI. This experiment aims to verify whether the proposed method can learn to discover the semantic parts under such challenging conditions. As shown in Table 2, without pretraining, the proposed method can still learn to discover semantic face parts, achieving comparable performance to GANSeg and outperforming other state-of-the-art methods with a large margin in NME. By utilizing the knowledge learned by DINO, we observe significant improvements of 15.40\%, 23.45\% and 34.94\% in NME, NMI and ARI respectively, compared to the model without pretraining.

The second experiment is \textit{cross}-dataset validation. This experiment aims to assess the generalization ability of the proposed method. We train the proposed method on the training set of CelebA-in-the-wild, while the training set of AFLW is solely used for calculating the transformation matrix. Compared to \textit{within}-dataset validation, the model with pretrained DINO maintains comparable performance in the setting of \textit{cross}-dataset validation and achieves the second best performance among all methods listed in Table 2. Therefore, the ability of our model to utilize knowledge from pretrained models significantly enhances its robustness in unsupervised part discovery.

\begin{table}[t!]
	\centering
	\resizebox{1.00\linewidth}{!}{
		\begin{tabular}{|m{2.7cm}<{\centering}|m{1.0cm}<{\centering}|m{1.5cm}<{\centering}|m{0.6cm}<{\centering}|m{0.6cm}<{\centering}|m{0.6cm}<{\centering}|m{0.6cm}<{\centering}|}
			\hline
			Method & Pretrain &Auxiliary Supervision & FG-NMI$\uparrow$ & FG-ARI$\uparrow$ & NMI$\uparrow$ & ARI$\uparrow$ \\ \hline
			SCOPS \cite{SCOPS} & Y & foreground & 30.7 & {\color{blue} \textbf{27.6}} & {\color{blue} \textbf{56.6}} & {\color{blue} \textbf{81.4}} \\
			Choudhury et al. \cite{NIPSPart} & Y & foreground & {\color{red} \textbf{44.8}} & {\color{red} \textbf{46.6}} & {\color{red} \textbf{68.1}} & {\color{red} \textbf{90.6}} \\ 
			Huang et al. \cite{HuangAttention} & Y & Attribute & 14.0 & 9.6 & 15.4 & 25.0 \\
			PDiscoNet \cite{PDiscoNet} & Y & Attribute & 14.0 & 7.7 & 18.9 & 26.5 \\
			GANSeg \cite{GANSeg} & N & N/A & 29.6 & 26.0 & 47.2 & 75.9 \\  \hline
			Ours & N & N/A & 13.3 & 10.5 & 35.3 & 61.1 \\
			Ours$^\star$ & Y & N/A & {\color{blue} \textbf{34.1}} & 26.0 & 55.5 & 79.3 \\ \hline
		\end{tabular}
	}
	\caption{Performance comparisons on DeepFashion in the setting of $K=4$. (FG-)NMI and (FG-)ARI metrics are reported. Key: [{\color{red} \textbf{Best}}, {\color{blue} \textbf{Second best}}, $^\star$=pretrained on ImageNet using DINO]}
	\label{Table4}
\end{table}

\begin{table}[t!]
	\centering
	\resizebox{1.00\linewidth}{!}{
		\begin{tabular}{|m{2.9cm}<{\centering}|m{1.1cm}<{\centering}|m{1.5cm}<{\centering}|m{0.9cm}<{\centering}|m{0.9cm}<{\centering}|}
			\hline
			Method & Pretrain & Auxiliary Supervision & NMI $\uparrow$ & ARI $\uparrow$ \\ \hline
			Huang et al. (K=8) \cite{HuangAttention} & Y & class & 5.88 & 1.53 \\
			DINO (K=8) \cite{amir2021deep} & Y & N/A &  19.17 & 7.59 \\
			PDiscoNet (K=8) \cite{PDiscoNet} & Y & class & {\color{blue} \textbf{27.13}} & 8.76 \\ \hline
			Ours & N & N/A & 22.13 & {\color{blue} \textbf{33.82}} \\
			Ours$^\star$ & Y & N/A & {\color{red} \textbf{27.85}} & {\color{red} \textbf{40.77}} \\ \hline
		\end{tabular}
	}
	\caption{Performance comparisons on PartImageNet. NMI and ARI metrics are reported. Key: [{\color{red} \textbf{Best}}, {\color{blue} \textbf{Second best}}, $^\star$=pretrained on ImageNet using DINO]}
	\label{Table5}
\end{table}

\textbf{CUB-2011}: we evaluate our method on the first three categories and the full set. The corresponding results are reported as CUB-1/2/3 and CUB-F in Table 3, and some visualized results are shown in Fig. 6. Without pretraining on large-scale datasets, existing methods cannot divide the foreground object into parts in a manner close to human perception. Consequently, these methods often fail to discern the orientations of nearly symmetric objects. As shown in 2-nd and 3-rd column in Fig. 6, the green masks always focus on one side, while the blue masks always focus on the other side. This limitation is also mentioned in \cite{LiuDis}. Nevertheless, the proposed two-networks architecture allows the use of a pretrained model to learn a more coherent part-specific attention. As a result, the model with pretraining successfully overcomes this limitation and achieves improvements of 105.45\% in NMI and 108.60\% in ARI, respectively, compared to the model without pretraining.

\begin{figure}[t!]
	\centering
	\includegraphics[width=\linewidth]{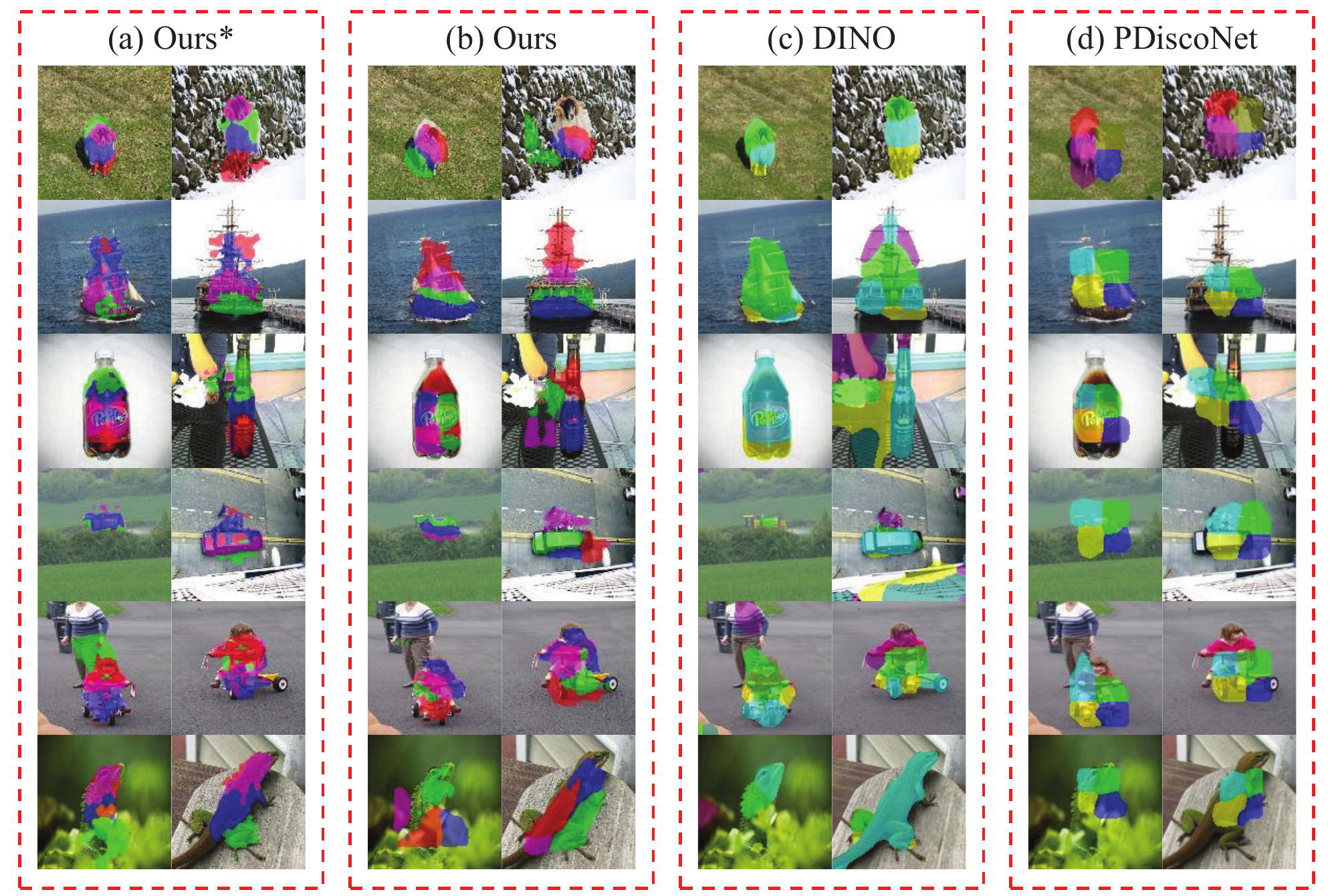}
	\caption{Visualized part discovery results of our proposed method and other state-of-the-art methods on PartImageNet. Key: [\textcolor[rgb]{1, 0, 0}{\textbf{Part1}}, \textcolor[rgb]{0, 1, 0}{\textbf{Part 2}}, \textcolor[rgb]{0, 0, 1}{\textbf{Part 3}}, \textcolor[rgb]{1, 0, 1}{\textbf{Part 4}}, \textcolor[rgb]{0, 1, 1}{\textbf{Part 5}}, \textcolor[rgb]{1, 1, 0}{\textbf{Part 6}}, \textcolor[rgb]{0.5, 0.5, 0}{\textbf{Part 7}}, \textcolor[rgb]{0.5, 0.0, 0.5}{\textbf{Part 8}}, $^\star$=pretrained on ImageNet using DINO]}
	\label{fig8}
\end{figure}

\begin{figure*}[t!]
	\centering
	\includegraphics[width=\linewidth]{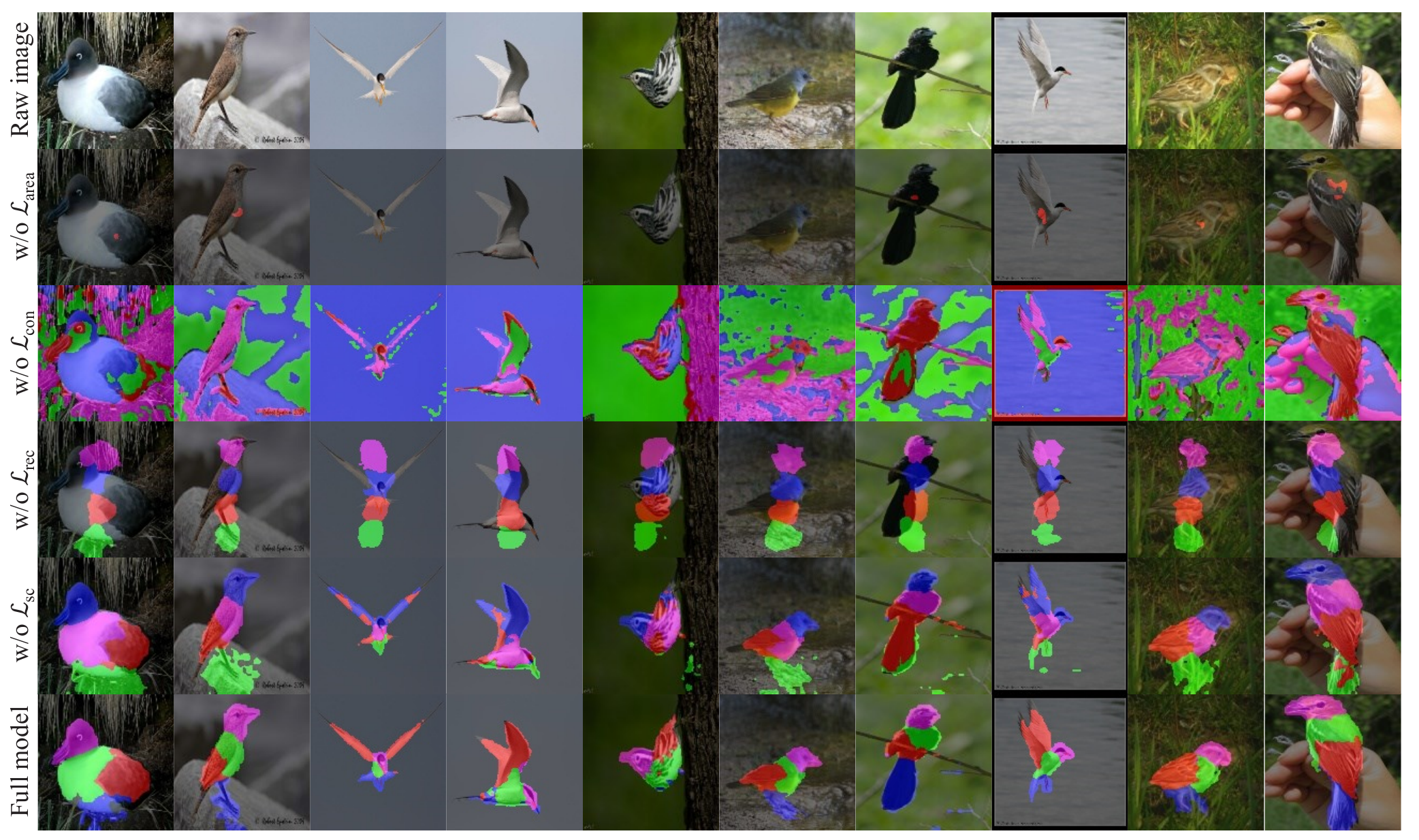}
	\caption{Ablation studies of loss functions. From top to bottom: 1) raw images, models trained without 2) area loss, 3) concentration loss, 4) reconstruction loss, 5) semantic consistency loss, 6) nothing (full model).}
	\label{fig9}
\end{figure*}

Without relying on any auxiliary information for supervision, the proposed method exhibits very competitive performance under both insufficient and sufficient training sample conditions. Remarkably, it even outperforms Choudhury et al., which is trained using foreground masks, in the metric of NMI. Moreover, the proposed method also outperforms Huang et al. and PDiscoNet, which discover object parts in a weakly-supervised manner. Similar to the visualized results on CelebA, the masks predicted by Huang et al. and PDiscoNet still contain rough part regions rather than explicit part boundaries. Compared to directly clustering the foreground pixels of DINO using K-means for part discovery and further utilizing the centers of these clusters as part representations as \cite{amir2021deep}, the part representations learned by the proposed Partfromer can adapt to different conditions. Consequently, with the same backbone, the proposed method improves the metrics by 16.20\% and 20.50\% in NMI and ARI respectively.

\begin{table}[t!]
	\centering
	\resizebox{1.00\linewidth}{!}{
		\begin{tabular}{|m{1.4cm}<{\centering}|m{1.4cm}<{\centering}|m{1.4cm}<{\centering}|m{1.4cm}<{\centering}|m{1.4cm}<{\centering}|m{1.4cm}<{\centering}|}
			\hline
			Method & w/o $\mathcal{L}_{\rm area}$ & w/o $\mathcal{L}_{\rm con}$ & w/o $\mathcal{L}_{\rm rec}$ & w/o $\mathcal{L}_{\rm sc}$ & Full model \\ \hline
			FG-NMI$\uparrow$& 0.0 & 4.4 & 12.0 & 41.8 & {\color{red} \textbf{44.8}} \\ \hline
			FG-ARI$\uparrow$& 0.0 & 1.0 & -3.7 & 18.2 & {\color{red} \textbf{19.3}} \\ \hline
			NMI$\uparrow$& 1.8 & 5.2 & 15.9 & 41.3 & {\color{red} \textbf{45.2}} \\ \hline
			ARI$\uparrow$& 1.0 & 1.1 & 4.7 & 17.7 & {\color{red} \textbf{19.4}} \\ \hline
		\end{tabular}
	}
	\caption{Performance comparison of the proposed method (with pretraining) without different loss functions on CUB-F dataset. Key: [{\color{red} \textbf{Best}}]}
	\label{Table6}
\end{table}

\textbf{DeepFashion}: as shown in Table 4, despite not employing any auxiliary labels for supervision, our method (with pretrained model) yields the second best performance in FG-ARI. This illustrates that the discovered parts have high consistency in semantics. Additionally, as shown in Fig. 7, the parts identified by our method are meaningful (namely: head \& hair, upper-blouse, skin \& lower-blouse and trousers), making the predicted results useful for downstream tasks. Because both SCOPS and Choudhury et al. utilize labeled foreground masks for supervision and operate at higher image resolutions during both training and testing, they significantly outperform the methods without using foreground masks in terms of the NMI and ARI metrics. Nevertheless, our method achieves performance comparable to that of SCOPS in completely unsupervised settings, as measured by the NMI and ARI metrics. Despite the use of attribute labels, PDiscoNet and Huang et al. do not demonstrate competitive performance on Deepfashion. The main reason is that the masks predicted by PDiscoNet and Huang et al. on Deepfashion only contain rough regions of corresponding parts, as the masks predicted on CelebA. Unlike CelebA and CUB, Deepfashion provides dense masks for NMI and ARI calculation. Therefore, this limitation is reflected on the metrics, resulting in lower NMI and ARI.

\textbf{PartImageNet}: to apply our method to PartImageNet, which contains multiple classes, we generate $K \times N_{\rm class} + 1$ ($N_{\rm class}$ is the number of classes) part embeddings, including $K \times N_{\rm class}$ for foreground parts and one for background. During training and testing, the model uses only the $K$ corresponding part embeddings ($K$ is set to 4 in this experiment) and the shared background embedding for the specific class. Other parameters of the network are reused. The quantitative comparison results are shown in Table 5. Similar to the results on other challenging benchmarks, such as CUB and Deepfashion, we also observe significant improvements of 25.85\% in NMI and 20.55\% in ARI, brought by using pretrained weights. Therefore, we can conclude that the ability to utilize pretrained weights enables the proposed method to maintain strong robustness under very challenging conditions.

\begin{figure*}[t!]
	\centering
	\includegraphics[width=\linewidth]{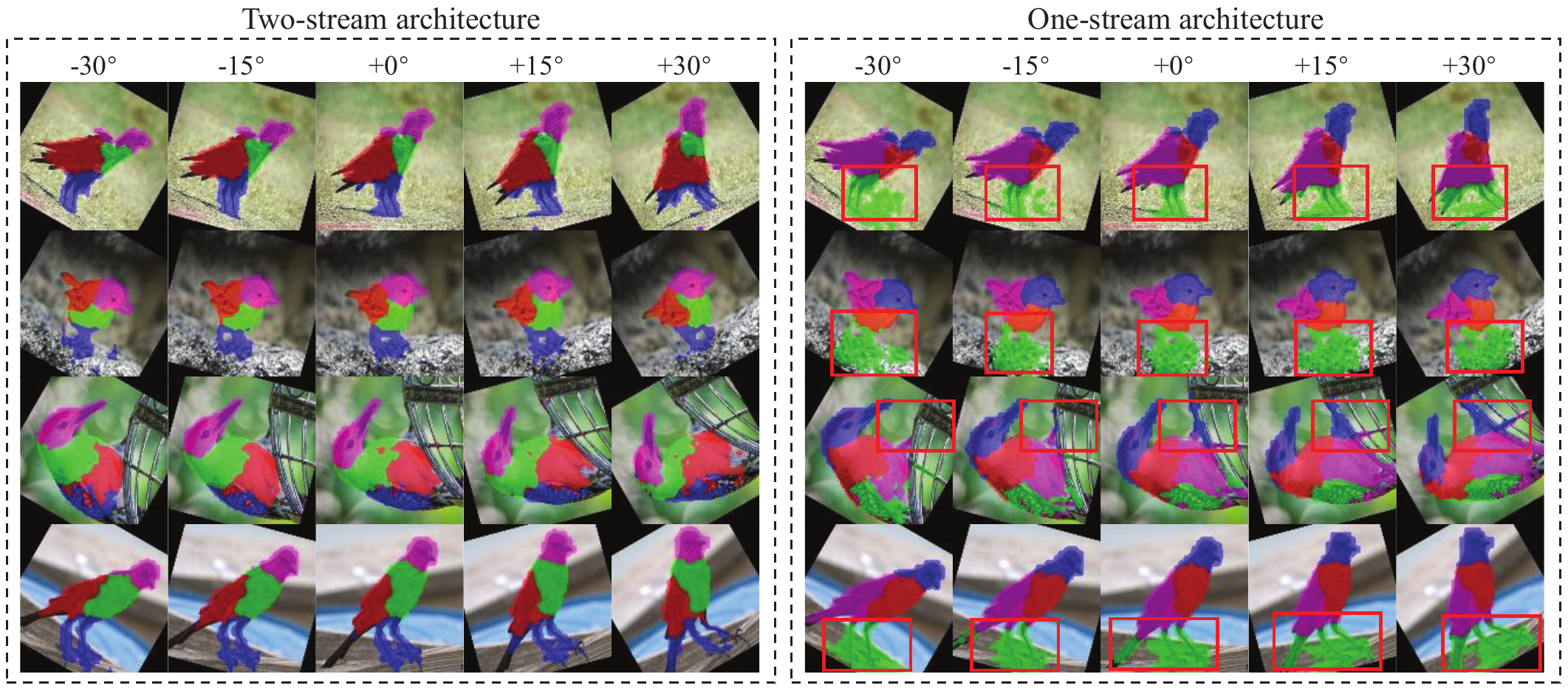}
	\caption{Visualization of some part discovery results on CUB-F across different rotation angles, predicted by models trained with and without part representation exchange.}
	\label{fig10}
\end{figure*}

\begin{table*}[t!]
	\centering
	\begin{tabular}{|m{1.8cm}<{\centering}|m{0.7cm}<{\centering}|m{0.7cm}<{\centering}|m{0.7cm}<{\centering}|m{0.7cm}<{\centering}|m{0.7cm}<{\centering}|m{0.7cm}<{\centering}|m{0.7cm}<{\centering}|m{0.7cm}<{\centering}|m{0.7cm}<{\centering}|m{0.7cm}<{\centering}|}
		\hline
		\multirow{2}{*}{Method} & \multicolumn{2}{c|}{-30$^\circ$} &\multicolumn{2}{c|}{-15$^\circ$} & \multicolumn{2}{c|}{+0$^\circ$} & \multicolumn{2}{c|}{+15$^\circ$} & \multicolumn{2}{c|}{+30$^\circ$} \\ \cline{2-11}
		& NMI$\uparrow$ & ARI$\uparrow$ & NMI$\uparrow$ & ARI$\uparrow$ & NMI$\uparrow$ & ARI$\uparrow$ & NMI$\uparrow$ & ARI$\uparrow$ & NMI$\uparrow$ & ARI$\uparrow$ \\ \hline
		One-stream & 39.0 & 16.8 & 42.1 & 18.0  & 43.6 & 18.4 & 42.0 & 17.9 & 38.5 & 16.6 \\ \hline
		Two-stream & {\color{red} \textbf{40.2}} & {\color{red} \textbf{17.4}} & {\color{red} \textbf{43.8}} & {\color{red} \textbf{18.8}} & {\color{red} \textbf{45.2}} & {\color{red} \textbf{19.4}} & {\color{red} \textbf{43.7}} & {\color{red} \textbf{18.7}} & {\color{red} \textbf{40.0}} & {\color{red} \textbf{17.3}} \\ \hline
	\end{tabular}
	
	\caption{Ablation study of the proposed method trained with/without exchanging part representations within paired images on CUB-F across different rotation angles. Key: [{\color{red} \textbf{Best}}]}
	\label{Table7}
\end{table*}

Despite the using of class labels in PDiscoNet, the visualized results predicted by PDiscoNet are not as good as the results predicted by DINO and our method, as shown in Fig. 8. Its predicted masks still contain only rough regions of corresponding parts, rather than clear part boundaries. As a result, our model with pretrained weight still outperforms PDiscoNet by 2.65\% and 315.41\% in the metric of NMI and ARI respectively. Compared to directly clustering the features of DINO as \cite{amir2021deep}, our proposed method adaptively extracts part representations from input images, rather than using the means of clusters as part representations. Therefore, the learned part representations can adapt to different conditions, leading to better semantic consistency of the discovered parts. As a result, we can observe improvements of 45.28\% and 437.15\% in the metric of NMI and ARI respectively, despite using the same backbone.

\subsection{Ablation Studies}

\textbf{Influence of loss functions}: to investigate the contributions of different loss functions, we train the models under different configurations of loss functions using the CUB-F dataset. The quantitative results are reported in Table 6, while some visualized results are presented in Fig. 9. The geometric constraints, area loss ($\mathcal{L}_{\rm area}$) and concentration loss ($\mathcal{L}_{\rm con}$), play an important role in learning the part-specific attention. $\mathcal{L}_{\rm con}$ is the key to dual representation alignment and local attention learning. Otherwise, the strong global receptive field of ViT drives itself to look at the entire image, resulting in the model degrading to a color cluster model. Without $\mathcal{L}_{\rm area}$, the model tends to satisfy $\mathcal{L}_{\rm con}$ as a priority. Consequently, the model converges to trivial solutions, identifying nearly all pixels as background.

The semantic constraints, reconstruction loss $\mathcal{L}_{\rm rec}$ and semantic consistency loss $\mathcal{L}_{\rm sc}$, are also crucial for object part discovery. $\mathcal{L}_{\rm rec}$ ensures the learned part representations explicitly contain features of the corresponding parts, rather than being a series of meaningless vectors. As shown in the 4-$th$ row of Fig. 9, without using $\mathcal{L}_{\rm rec}$ to align the dual representations, the segmentation results also become meaningless and cannot follow the changes in object position and pose. $\mathcal{L}_{\rm sc}$ enlarges the cosine distance between the representations of different parts, ensuring the consistency of the discovered parts. As shown in 3-$rd$ and 4-$th$ columns of the 5-$th$ row in Fig. 9, the wings are simultaneously segmented into two different parts due to the lack of $\mathcal{L}_{\rm sc}$. The absence of this constraint also results in the model segmenting some background pixels, whose semantics are significantly inconsistent with the target, as foreground parts. The influence of $\mathcal{L}_{\rm sc}$ on the part representation distribution is further discussed in the supplementary file.

\textbf{Influence of part representation exchange within paired images}: we also implement a model without exchanging part representations within the paired images, using $\bm{G}$ and $\bm{F}$ from the same image for reconstruction. To investigate the influence of part representation exchange during training on geometric transformation invariance, we rotate the images from CUB-F by a certain angle and input them into both models trained with/without part representation exchange for comparison. The quantitative results are tabulated in Table 7, and some qualitative results are shown in Fig. 10. As shown in Fig. 10, the semantics of the results predicted by the the model trained with part representation exchange are more consistent than those predicted by the model without exchange. The model lacking part representation exchange may incorrectly segment the background as foreground parts when the input images are rotated by a large angle, leading to significant performance degradation.  Consequently, the model with part representation exchange outperforms the one without by a noticeable margin in both NMI and ARI across all angles.

\textbf{Visualization of part representations exchange across unpaired images}: to further analyze the specific content contained in part representations, we exchange the foreground part representations from two different input images and use these exchanged representations for reconstruction. The reconstructed images are shown in Fig. 11. We could observe that most facial components of the two input images are exchanged in the reconstructed images. This illustrates that the part representations explicitly contain the features of corresponding parts, including shape and color, which is why the part representations can serve as a series of reliable detectors in object part discovery. However, we also notice that some facial features have leaked from the original image. As shown in the probability maps $\bm{V}$ of Fig. 2, the background confidence value is not zero near the edges of foreground parts. Therefore, the representations of synthetic feature map $\bm{S}$ in these regions still contain a small proportion of the background representation. Using the synthetic feature map $\bm{S}$ for reconstruction results in some facial features being recorded in the background representation. Therefore, the generated faces retain some features from the original faces.

\begin{figure}[t!]
	\centering
	\includegraphics[width=\linewidth]{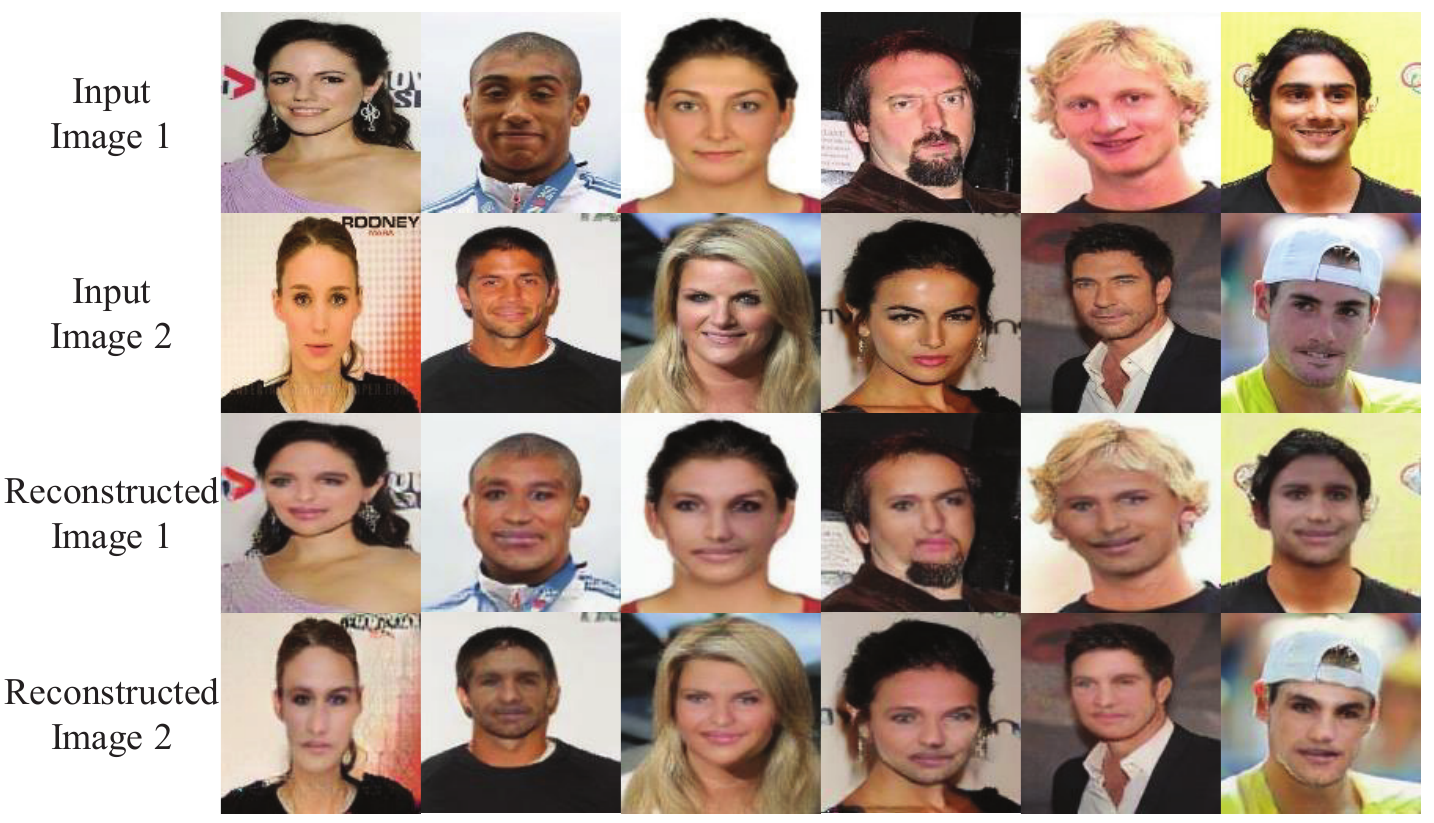}
	\caption{Some reconstructed faces using the exchanged part representations among unpaired faces}
	\label{fig11}
\end{figure}

\textbf{Influence of different pretraining methods}: to further explore the influence of different pretrain methods on part discovery, we carry out ablation studies on CelebA-in-the-wild using \textbf{frozen ViT-B/16s} \cite{VIT} pretrained with three widely used methods: DINO \cite{DINO}, MAE \cite{MAE} and MoCo V3 \cite{MoCoV3}, and a ViT-B/16 initialized from scratch. The quantitative results are reported in Table 8 and some qualitative results are shown in Fig. 12. MoCo V3 cannot produce feature maps with significant part-level features. As a result, the proposed method cannot achieve part discovery using its pretrained weight. Both DINO and MAE produce highly similar features in face regions. The models with these two pretrained models cannot divide the face region into the target number of parts and tend to identify other regions with less semantic consistency, such as hair, neck and shoulders, as foreground parts. In contrast, the model initialized from scratch easily divides the face region into the target number of parts. The primary reason is that the human face appears in ImageNet \cite{ImageNet} as part of a person rather than as an independent target. It is hard for these pretrained models to further decompose the human face, a part of a person, into the target number of parts and produce finer-grained face representation. This also explains why the attention mechanism of DINO tends to view the face regions as background and focus on other objects, such as clothes and sunglasses, as shown in Fig. 4. Therefore, we do not observe an improvement on CelebA-in-the-wild as significant as the improvement on other datasets brought by DINO.

\begin{figure}[t!]
	\centering
	\includegraphics[width=0.9\linewidth]{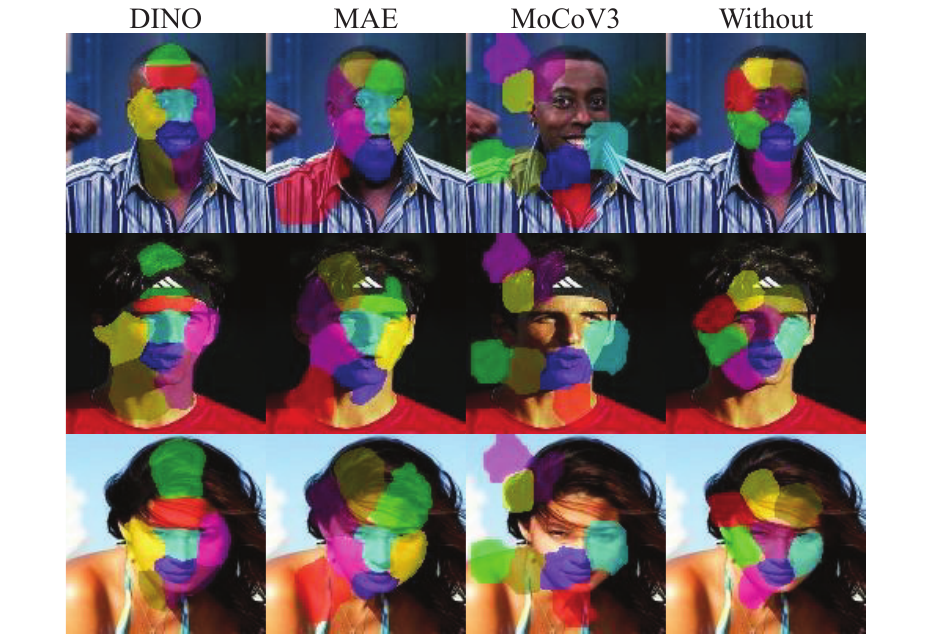}
	\caption{Visualization of part discovery results predicted by the proposed method with different pretrained weights on CelebA-in-the-wild (K=8).}
	\label{fig12}
\end{figure}

\textbf{Influence of $\alpha$}: $\alpha$ serves as important prior knowledge to control the expected minimum size of the predicted part area. To investigate the influence of $\alpha$, we train our proposed method under different $\alpha$ settings. As shown in Table 9, although the selection of $\alpha$ is crucial for the training of our method, the method still outperforms previous unsupervised part discovery methods across a wide range of $\alpha$ values. This demonstrates that the proposed method is relatively robust to variations in hyperparameters.

% When $\lambda_{\rm sc}$ is set within $[0, 0.01]$, with the increasing of $\lambda_{\rm sc}$, the performance of the proposed method is also improved. 

\begin{table}[t!]
	\centering
	\resizebox{1.00\linewidth}{!}{
		\begin{tabular}{|m{1.1cm}<{\centering}|m{1.7cm}<{\centering}|m{1.7cm}<{\centering}|m{1.7cm}<{\centering}|m{1.7cm}<{\centering}|}
			\hline
			Pretrain & DINO\cite{DINO} & MAE\cite{MAE} & MoCoV3\cite{MoCoV3} & Without \\ \hline
			NMI$\uparrow$& 43.7 & {\color{red} \textbf{59.5}} & 22.0 & 50.5 \\ \hline
			ARI$\uparrow$& 28.1 & {\color{red} \textbf{41.2}} & 7.4 & 38.0 \\ \hline
			NME$\downarrow$& 7.60 & {\color{red} \textbf{7.45}} & 47.6 & 7.77 \\ \hline
		\end{tabular}
	}
	\caption{Performance comparison of the proposed method on CelebA-in-the-wild ($K=8$) using different pretraining methods. Key: [{\color{red} \textbf{Best}}]}
	\label{Table8}
\end{table}

\begin{table}[t!]
	\centering
	\begin{tabular}{|m{1.3cm}<{\centering}|m{0.6cm}<{\centering}|m{0.6cm}<{\centering}|m{0.6cm}<{\centering}|m{0.6cm}<{\centering}|m{0.6cm}<{\centering}|}
		\hline
		Method & $\alpha$=0.1 & $\alpha$=0.3 & $\alpha$=0.5 & $\alpha$=0.7 & $\alpha$=0.9 \\ \hline
		FG-NMI$\uparrow$& 36.7 & 44.4 & {\color{red} \textbf{44.8}} & 41.6 & 41.4 \\ \hline
		FG-ARI$\uparrow$& 14.6 & 17.8 & {\color{red} \textbf{19.3}} & 17.8 & 17.1 \\ \hline
		NMI$\uparrow$& 44.9 & {\color{red} \textbf{48.1}} & 45.2 & 42.8 & 42.0 \\ \hline
		ARI$\uparrow$& {\color{red} \textbf{22.0}} & 20.9 & 19.4 & 18.6 & 17.3 \\ \hline
	\end{tabular}
	\caption{Performance comparison of the proposed method on CUB-F in the setting of different $\alpha$. Key: [{\color{red} \textbf{Best}}]}
	\label{Table9}
\end{table}

\textbf{Influence of different reconstruction losses}: to study the influence of two commonly used reconstruction losses, MSE loss and perceptual loss, we implement two models on the CelebA-in-the-wild: one trained with MSE loss and the other trained with perceptual loss. The quantitative results are presented in Table 13, and some parts discovery results predicted by the model with MSE loss are shown in Fig. 10. Additionally, we also exchange the foreground part representations from two different images and reconstruct the images using the decoder. As shown in Fig. 13, the reconstructed images are blurred and lack details in the facial components. Exchanging part representations only leads to a change in color in specific regions, rather than a change in appearance. This indicates that the model trained with MSE loss fails to learn the semantics of the corresponding object parts. As a result, the part representations learned with MSE loss cannot serve as a series of reliable detectors for segmenting the corresponding parts from targets during the testing phase. Similar to the model trained without a reconstruction loss, the model trained with MSE loss is unable to accurately segment part boundaries from images, leading to significant performance degradation in NME.

\begin{figure}[t!]
	\centering
	\includegraphics[width=\linewidth]{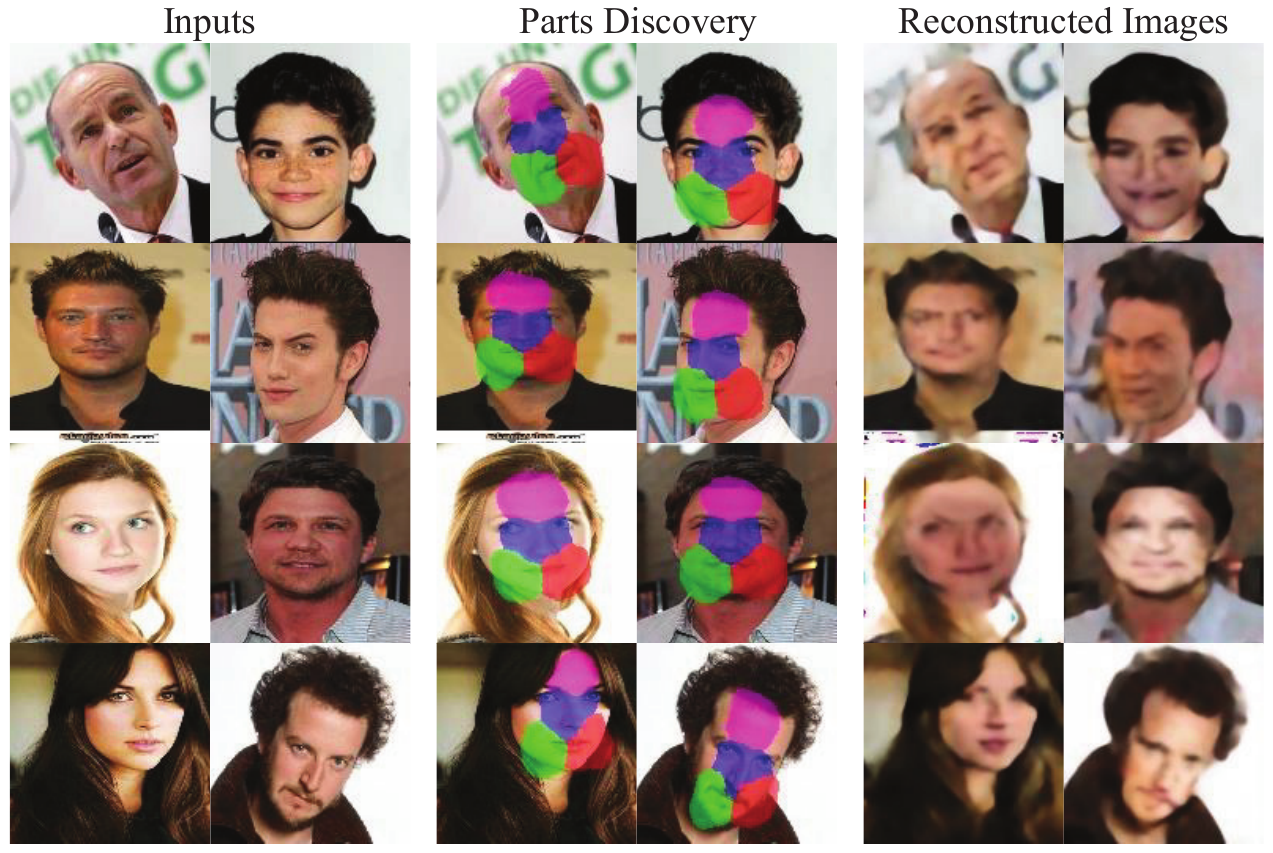}
	\caption{The input images (1st and 2nd column), parts discovery results (3rd and 4th column) and reconstructed images with part representation exchange (5th and 6th column) of the model trained with \textbf{MSE} loss.}
	\label{fig13}
\end{figure}

\begin{table}[t!]
	\centering
	\begin{tabular}{|m{2.4cm}<{\centering}|m{1.6cm}<{\centering}|m{1.6cm}<{\centering}|}
		\hline
		\multirow{2}{*}{Interpolation} & \multicolumn{2}{c|}{Inter-ocular NME $\downarrow$} \\ \cline{2-3}
		& K=4 & K=8 \\ \hline
		MSE Loss & 16.83\% & 16.70\% \\ \hline
		Perceptual Loss & {\color{red} \textbf{13.28\%}} & {\color{red} \textbf{8.70\%}} \\ \hline
	\end{tabular}
	\caption{Performance comparison of the proposed method with perceptual or MSE losses on CelebA-in-the-wild (without pretraining). Key: [{\color{red} \textbf{Best}}]}
	\label{Table10}
\end{table}

\textbf{Influence of feature map interpolation}: to investigate the influence of feature map interpolation during the testing phase, we implement two different versions: one with interpolation and one without interpolation. The model weights remain the same in these two versions. The quantitative results on CUB-F dataset are presented in Table 11 and some qualitative results are shown in Fig. 14. Despite using the same training weights, we observe that the version with interpolation produces results with tighter and smoother boundaries. Consequently, the interpolation process improves the metrics by 6.10\% and 3.19\% in NMI and ARI, respectively.

\subsection{Limitations and future work}

We demonstrate some failure cases predicted by our method on PartImageNet in Fig. 15. If the number of the images containing multiple objects (fish and human) exceeds a certain ratio in a dataset, the proposed method may mistakenly identify the non-target object (human) as the foreground. Consequently, the target object (fish) cannot be properly decomposed into $K$ parts, which further results in failures in part discovery. Nevertheless, images containing multiple objects will inevitably be introduced into training. In the future, we will make efforts to address this limitation by utilizing the prior knowledge contained in open-vocabulary.

\begin{figure}[t!]
	\centering
	\includegraphics[width=\linewidth]{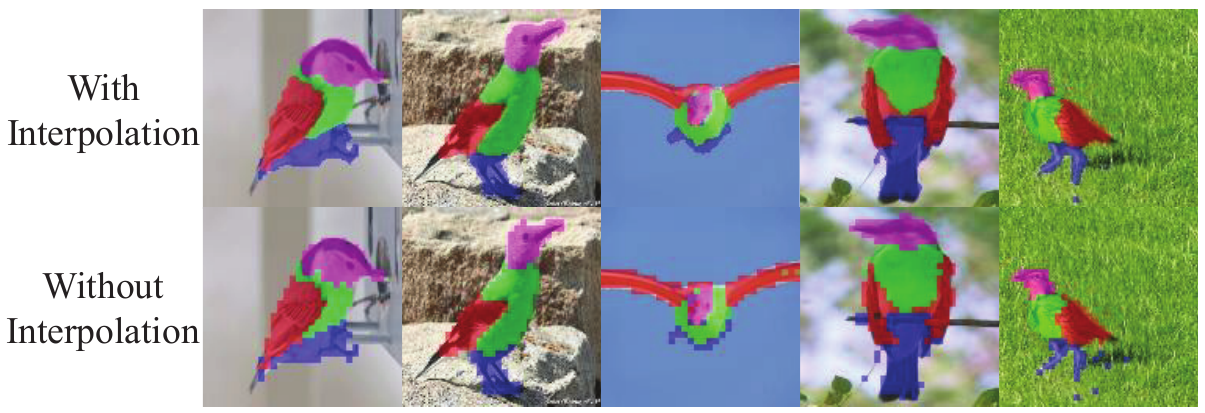}
	\caption{Some visualized qualitative results predicted with/without feature map interpolation on CUB-F.}
	\label{fig14}
\end{figure}

\begin{figure}[t!]
	\centering
	\includegraphics[width=0.6\linewidth]{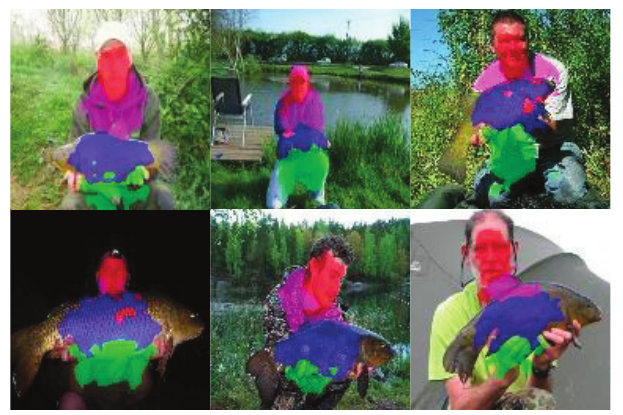}
	\caption{Some failure cases predicted by the proposed method on PartImageNet}
	\label{fig15}
\end{figure}

\begin{table}[t!]
	\centering
	%\resizebox{1.00\linewidth}{!}{
	\begin{tabular}{|m{1.8cm}<{\centering}|m{1.3cm}<{\centering}|m{1.3cm}<{\centering}|m{1.0cm}<{\centering}|m{1.0cm}<{\centering}|}
		\hline
		interpolation & FG-NMI$\uparrow$ & FG-ARI$\uparrow$ & NMI$\uparrow$ & ARI$\uparrow$ \\ \hline
		without  & 42.5 & 18.6 & 42.6 & 18.8 \\ \hline
		with & {\color{red} \textbf{44.8}} & {\color{red} \textbf{19.3}} & {\color{red} \textbf{45.2}} & {\color{red} \textbf{19.4}} \\ \hline
	\end{tabular}
	%}
	\caption{Performance comparisons of the proposed method with/without feature map interpolation in testing phase on CUB-F (with pretraining). Key: [{\color{red} \textbf{Best}}]}
	\label{Table11}
\end{table}

\section{Conclusion}
We propose a novel paradigm for unsupervised part-specific attention learning. This approach applies semantic and geometric constraints to the proposed PartFormer based on dual representation alignment, successfully driving the learned representations to explicitly capture the information of the corresponding parts. The dual representation alignment also evolves these part representations into a series of reliable detectors for unsupervised part discovery. Incorporating with the part-specific attention, the discovered parts demonstrate consistent semantics, even in complex backgrounds and extreme conditions. Additionally, the novel two-network design enables the use of pretrained weights. By leveraging knowledge gained from pretraining on large-scale datasets, our method can learn part-specific attention that more closely approximates human perception. Consequently, it successfully addresses a common limitation of previous works: the inability to discern the orientations of nearly symmetric objects. Extensive experiments on five widely-used benchmarks demonstrate the highly competitive performance of our method, and ablation studies illustrate the effectiveness of the constraints incorporated into the architecture. In the future, we plan to extend our proposed method to large-scale datasets with an open-vocabulary. By aligning part representations with textual descriptions, the model can learn more coherent part-specific attention and produce finer-grained pixel-level representations for these classes.

% Can use something like this to put references on a page
% by themselves when using endfloat and the captionsoff option.
\ifCLASSOPTIONcaptionsoff
  \newpage
\fi

\bibliographystyle{IEEEtran}
\bibliography{egbib.bib}

% that's all folks
\end{document}